\documentclass{article}

\usepackage{xr}
\makeatletter

\newcommand*{\addFileDependency}[1]{
\typeout{(#1)}
\@addtofilelist{#1}
\IfFileExists{#1}{}{\typeout{No file #1.}}
}\makeatother

\makeatletter
\def\@email#1#2{%
 \endgroup
 \patchcmd{\titleblock@produce}
  {\frontmatter@RRAPformat}
  {\frontmatter@RRAPformat{\produce@RRAP{*#1\href{mailto:#2}{#2}}}\frontmatter@RRAPformat}
  {}{}
}%
\makeatother

\usepackage{arxiv,times}
\usepackage[authoryear]{natbib}
\usepackage{graphics, graphicx, placeins}
\graphicspath{ {./media/} }
\usepackage{algorithm}
\usepackage{algpseudocode}
\usepackage{dcolumn}  
\usepackage[T1]{fontenc}
\usepackage{color} 
\usepackage{bm} 
\usepackage[dvipsnames]{xcolor,colortbl}
\usepackage{amsmath, amssymb, amsfonts, etoolbox}
\usepackage{hyperref}
\usepackage{cleveref}
\usepackage{tabularx}
\usepackage{longtable}
\usepackage{xltabular}

\usepackage{xspace}
\newcommand{\ie}{\textit{i.e.}\xspace}
\newcommand{\eg}{\textit{e.g.}\xspace}
\newcommand{\cf}{\textit{c.f.}\xspace}

\newcommand{\insilico}{\textit{in silico}\xspace}

\def\HADES{\textit{HADES}\xspace}
\def\CHARLES{\textit{CHARLES-D}\xspace}
\def\popsize{{N_p}}
\def\sigmainit{{\sigma_I}}
\def\genome{{\bm{g}}}            
\def\population{{\bm{G}}}        
\def\dataset{{\bm{X}}}        
\def\condition{{\bm{c}}}         
\def\target{{\mathrm T}}
\def\ceval{c}                      
\def\gt{{\tau}}
\def\vx{{\bm x}}
\def\ve{{\bm\epsilon}}

\usepackage{epstopdf}
\AppendGraphicsExtensions{.tiff}

\usepackage[normalem]{ulem}       
\usepackage{blindtext}

\newcommand{\ccharles}[1]{\textcolor{olive}{#1}}
\newcommand{\tcharles}[1]{\textcolor{teal}{#1}}

\newcommand{\params}[1]{\textcolor{black}{#1}}  

\title{Heuristically Adaptive Diffusion-Model Evolutionary Strategy}
    
\author{Benedikt Hartl$^{1,2*}$ \quad Yanbo Zhang$^{1*}$ \quad Hananel Hazan$^{1\dagger}$\thanks{Equal contributions. $^\dagger$ Author of correspondence: \texttt{Hananel@Hazan.org.il}} \quad Michael Levin$^{1,3}$\\
$^1$ Allen Discovery Center at Tufts University, Medford, MA, 02155, USA\\
$^2$ Institute for Theoretical Physics, TU Wien, Austria\\
$^3$ Wyss Institute for Biologically Inspired Engineering at Harvard University,\\
~~~Boston, MA, 02115, USA
}

\begin{document} 
\maketitle

\begin{abstract}
Diffusion Models represent a significant advancement in generative modeling, employing a dual-phase process that first degrades domain-specific information via Gaussian noise and restores it through a trainable model. This framework enables pure noise-to-data generation and modular reconstruction of, \textit{e.g.}\xspace, images or videos. Concurrently, evolutionary algorithms employ optimization methods inspired by biological principles to refine sets of numerical parameters encoding potential solutions to rugged objective functions. Our research reveals a fundamental connection between diffusion models and evolutionary algorithms through their shared underlying generative mechanisms: both methods generate high-quality samples via iterative refinement on random initial distributions. By employing deep learning-based diffusion models as generative models across diverse evolutionary tasks and iteratively refining diffusion models with heuristically acquired databases, we can iteratively sample potentially better-adapted offspring parameters, integrating them into successive generations of the diffusion model. This approach achieves efficient convergence toward high-fitness parameters while maintaining explorative diversity. Diffusion models introduce enhanced memory capabilities into evolutionary algorithms, retaining historical information across generations and leveraging subtle data correlations to generate refined samples. We elevate evolutionary algorithms from procedures with shallow heuristics to frameworks with deep memory. By deploying classifier-free guidance for conditional sampling at the parameter level, we achieve precise control over evolutionary search dynamics to further specific genotypical, phenotypical, or population-wide traits. Our framework marks a major heuristic and algorithmic transition, offering increased flexibility, precision, and control in evolutionary optimization processes.
\end{abstract}
\maketitle
\section{Introduction}
\label{sec:introduction}
Two fundamental mechanisms in the biosphere are known to drive novelty: evolution and learning. Conventionally, evolution is understood as a gradual, slow variational process adapting organisms or lineages across generations to changing environmental conditions through natural selection~\citep{darwin1959origin, dawkins2016selfish}. In contrast, learning is a rapid transformational process enabling individuals to acquire knowledge and generalize based on subjective experiences within their lifetime~\citep{kandel_principles_2013, courville2006bayesian, holland2000emergence, dayan_theoretical_2001}. These mechanisms are extensively researched in separate domains of artificial intelligence, and recent studies have begun highlighting similarities between evolution and learning~\citep{watson2023collective, vanchurin2022toward, levin2022technological, watson2022design, kouvaris2017evolution, watson2016can, watson2016evolutionary, power2015can, hinton1987learning, baldwin2018new}.

Evolutionary algorithms (EAs) implement a search process using biologically inspired variational principles to iteratively refine sets of numerical parameters that encode potential solutions to often rugged, custom objective functions~\citep{vikhar2016evolutionary, goldberg1989genetic, grefenstette1993genetic, holland1992adaptation}. 
Traditional black box EAs utilize heuristic population data and associated fitness scores (\ie, evaluations of the objective function reminiscent to the fitness of a phenotype in its environment) to sample new, potentially better adapted candidate solutions (\ie, offspring comprising the next generation).
Depending on the specific implementation of the EA~\citep{Katoch2020}, this sampling process of novel genotypic parameters can be population-based through recombination and mutation operations at the genotypic level of thereby iteratively adapted generations, or even leveraged by sampling novel data-points from successively re-parameterized probabilistic models, \eg, with a Gaussian prior~\citep{hansen2001completely}.

Evolutionary algorithms incorporating generative processes operate on partial, heuristically-derived knowledge of the fitness landscape. These algorithms iteratively train probabilistic models using progressively refined data to identify and exploit genotypic correlations across generations. Through continuous refinement of the generative model, they aim to increase the probability of sampling high-fitness solutions, potentially accelerating optimization.
While this approach offers computational advantages, it presents notable challenges. The use of parameterized models to learn the manifold of correlated genotypic parameters introduces an inductive bias that may constrain the evolutionary search. This learned structure could limit exploration of novel solutions and lead to premature convergence, potentially leaving promising regions of the search space unexplored.

Moreover, it has become increasingly recognized that the evolution process does not simply build phenotypes adapted to specific environmental conditions~\citep{Buckley2024NaturalInduction,McMillen2024Collective,levin2023darwin,levin2022technological}: While this certainly is a by-product, evolution seems to primarily bring forth agential problem solving machines on all scales in the biosphere. 
While selection occurs at the phenotypic scale, most organisms are originally compressed into a single fertile cell. 
Following recent insights from developmental evolutionary biology~\citep{mitchell2024genomic, levin2023darwin,pezzulo_2015_remembering, pezzulo_2016_topdown}, the genome does not represent a direct blueprint of all details of the mature organism, it rather instantiates a generative model to construct the latter: genes encode protein sequences molecular hardware that  enables cells and cellular collectives to communicate, compute, and navigate the space of anatomical possibilities (morphospace) in ways that can reach their target morphology, or some other functional embodiment, despite novel scenarios and perturbations~\citep{LevinCollectiveIntelligence} .
Through hierarchical steps of development, cells proliferate and organize into tissue, vessels, organs, bones, etc., until the organism is constructed and refined during morphogenesis. 
Thus, there is an entire layer of physiological computation, which can perform context-sensitive problem-solving, between an organism's genes and its anatomical (phenotypic) form and function, rendering the relationship between genes and phenotypic traits not only indirect but in principle computationally irreducible~\citep{Wolfram2002} and emergent~\citep{levin2023darwin}.
This self-orchestrated developmental process exhibits fundamental plasticity in both structure and function, enabling the substrate comprising an organism to adapt to novel internal and environmental stressors~\citep{levin2023darwin}. Such adaptive capability can be understood as a form of collective intelligence~\citep{McMillen2024Collective} aligning with William James' definition~\citep{Fields2023Regulative}: ``Intelligence is the ability to reach the same goal by different means''.
Our biosphere is organized as a multi-scale competency architecture~\citep{levin2023darwin}, which in turn has dramatic implications on the underlying evolutionary process~\citep{hartl2024evolutionary, Shreesha2023}, which, however, is hardly aligned with current evolutionary search strategies.

This ability to adaptively respond to challenges and solve problems is not limited to the scale of cells and organisms in behavioral and morphogenetic spaces. At larger levels of organization, evolutionary dynamics demonstrate characteristics of adaptive associative memory, operating at both individual lineage~\citep{Watson2014} and ecosystem levels~\citep{power2015can}. This memory-like behavior manifests through Hebbian learning mechanisms~\citep{hebb1949organization}, that exhibit a similar dynamic of  to associative memory in neural networks~\citep{Hopfield1982}.

Moreover, evolution and adaptation can occur in leaps~\citep{szathmary2015toward}, responding rapidly to changing environmental conditions, sometimes within just a few generations~\citep{GarciaCastillo2024}. This open-ended process appears to inherently promote diversity and adaptability~\citep{stanley2015greatness, lehman2011abandoning, pugh2016qualitydiversity} , though the underlying mechanisms remain incompletely understood.

Recent approaches have begun bridging the conceptual gaps between evolutionary biology, developmental biology, and technological applications. One promising direction employs Variational Auto-Encoders (VAEs)~\citep{kingma2013auto} to create low-dimensional genotypic search spaces for evolutionary algorithms, while conducting fitness evaluations in higher dimensional parameter spaces~\citep{gaier2020discovering, bentley2022evolving}. A related approach, termed Deep Optimisation~\citep{Caldwell2022DeepOpt}, leverages multi-level evolutionary transitions, modeled with deep learning, to solve complex combinatorial problems with polynomial scaling. However, these methods tend to exhibit greedy behavior and require careful curation of the training dataset and architecture to enable exploration in a successively refined latent space representation rather than being constrained by the decoder's canalizing output~\citep{Caldwell2022DeepOpt}. While they effectively compress information from the training set to enable generalization, they do so by prioritizing a low bias–variance trade-off that minimizes variance on training error, ultimately limiting their capacity to generalize beyond the training data.
Significant questions remain about how biological evolution generates novelty~\citep{levin2023darwin, watson2022design, watson2016evolutionary, Frank2019, Frank2018, Frank2013, Frank2007, Wagner2014, Wagner2011, Tusscher2011, Wagner1996}. Computational techniques such as novelty search~\citep{lehman2011abandoning} and quality-diversity algorithms~\citep{pugh2016qualitydiversity} have made progress in addressing these questions, though gaps in our understanding persist.
Recent efforts have started drawing connections between evolutionary processes and broader concepts of intelligence~\citep{mitchell2024genomic,levin2023darwin,pezzulo_2016_topdown} and learning theory~\citep{kouvaris2017evolution, watson2016can, watson2016evolutionary, power2015can, Watson2014}.

Neural Cellular Automata (NCAs)~\citep{hartl2024evolutionary,Mordvintsev2022,Mordvintsev2020,Li2002}, computational tools from the field of \textit{Artificial Life}~\citep{Langton1997}, serve as generative models for studying morphogenesis \insilico through collective multi-agent processes. NCAs are particularly valuable for investigating the indirect encoding relationship between genotype and phenotype, as they capture the emergent, multi-scale properties inherent in biological development. Our recent work~\citep{hartl2024evolutionary} demonstrates that hierarchical functional genotypic encoding—a characteristic found in biological systems and modeled by NCAs—fundamentally influences evolutionary processes and enables rapid, modular adaptation to environmental changes~\citep{Wagner2007, Schlosser2004, Calabretta2003, Tusscher2011, Wagner1996}.
The multi-scale competency architecture of NCAs manifests through local cell-cell interactions generating global patterns and behaviors. This architecture enables collective pattern formation through distributed computation, while maintaining robust development despite perturbations. Furthermore, it facilitates adaptive responses across multiple spatial and temporal scales, culminating in the self-organization of modular, hierarchical structures. These characteristics mirror biological development's collective (intercellular) intelligence. However, training NCAs through gradient or evolutionary methods remains challenging.
Related promising approaches such as neuro-evolution techniques~\citep{Stanley2002}, growing and self-assembling artificial neural networks~\citep{najarro2022hypernca}, neural developmental programming~\citep{najarro2023towards}, and self-modeling approaches~\citep{chang2018neural, premakumar2024unexpectedbenefitsselfmodelingneural, zagal2011towards} have further expanded our understanding. These methodologies collectively enhance our comprehension of how indirect encoding facilitates the emergence of complex, adaptive behaviors from simple, local rules.

In contrast, recent breakthroughs in generative deep learning, particularly through diffusion models (DMs)~\citep{Dhariwal2021DMBeatGANs, ho2020denoising, song2020denoising, sohl2015deep}, have to significant advances in artificial intelligence. These models utilize stepwise autoregressive denoising to generate novel, realistic data points that conform  to complex target data distributions. Prominent implementations such as Stable Diffusion~\citep{rombach2022high} and Sora~\citep{brooks2024video} demonstrate unprecedented capabilities in generating diverse image and video content through conditional text prompts~\citep{rombach2022high, ho2022classifier}.
The impact of DMs extends significantly beyond visual content generation. In the field of computational biology, these models have advanced protein folding prediction~\citep{Jumper2021}. Within optimization domains, DMs enable generative multi-objective optimization through pretraining on closed datasets~\citep{yan2024EmoDM, Higham2023, krishnamoorthy2023diffusionmodelsblackboxoptimization, peebles2022learninglearngenerativemodels}. Moreover, their application to generative game play through world modeling~\citep{alonso2024diffusionworldmodelingvisual} demonstrates their versatility as a fundamental architecture for complex generative tasks across diverse domains.

Generative DMs iteratively transform Gaussian distributions into structured data-points that conform to the training data distribution~\citep{zhang2024diffevo}. Drawing inspiration from dissipative systems of non-equilibrium physics~\citep{sohl2015deep,Callen1951}, these models implement a two-phase process: first, a forward (diffusion) process progressively corrupts data points with incremental noise; then, a model learns the reverse (generative) process to predict and remove this noise. This noise prediction enables autoregressive gradually denoising during the generative process, where initially noisy samples are systematically refined until they match the statistical properties of the training data. The iterative denoising steps create a smooth trajectory through the latent space, allowing the model to capture complex, multi-modal data distributions with remarkable fidelity.

A key strength of diffusion models lies in their conditional training and sampling capabilities~\citep{rombach2022high, ho2022classifier}, providing precise control over the generative output and thus offering unprecedented flexibility in directing the generation process toward desired outcomes.
In general, DMs are tightly related to associative memory systems~\citep{Ambrogioni2023}, which fundamentally enables these generative models to sample diverse yet high-quality results on custom datasets while maintaining reliable training convergence.

The step-wise error correction process inherent to diffusion models bears striking resemblance to the mechanisms observed in NCAs, particularly in their application to modeling multicellular growth and development~\citep{hartl2024evolutionary}. This parallel suggests that diffusion models may be particularly well-suited for simulating computationally irreducible~\citep{Wolfram2002} self-orchestrated biological processes, including morphogenesis and evolution. The fundamental similarity in iterative refinement between diffusion models and biological development indicates shared underlying principles governing both artificial and natural generative processes, potentially offering new insights into the nature of generative systems across different domains. One important direction, to which our research contributes, is the detection of unexpected emergent dynamics in processes driven by apparently simple rules: not just complexity and unpredictability, but new forms of problem-solving (intelligence) can emerge in surprising places. It is essential to identify the underlying principles of such capabilities, in different embodiments (biological, technological, etc.) in order to advance engineering, mitigate risk of emerging technologies, and create a deeper understanding of biology and computation.

In our complementary work~\citep{zhang2024diffevo}, we established the mathematical equivalence between successive adaptation in evolutionary processes and the generative mechanisms of probabilistic diffusion models. This equivalence emerges from a fundamental similarity in their iterative refinement processes: evolutionary systems combine directed selection with random mutations, while diffusion models balance progressive denoising with stochastic perturbations.
In evolutionary processes, natural selection and genetic recombination guide populations toward higher fitness, while random mutations and locality maintain diversity. Similarly, diffusion models employ iterative denoising to transform random samples into meaningful data, incorporating controlled stochastic elements to ensure generative diversity. Building on these parallels, we introduce in Ref.~\citenum{zhang2024diffevo} the concept of model-free \textit{Diffusion Evolution} — a framework capable of efficiently exploring complex high-dimensional parameter spaces to generate diverse solutions for evolutionary optimization tasks. This approach achieves performance competitive with mainstream methods, notably without deploying deep learning architectures.
This unified perspective illuminates a broader theoretical connection: diffusion models can serve as a crucial bridge between evolutionary and developmental biology concepts and contemporary machine learning approaches. Such integration offers novel pathways for incorporating biological principles into artificial intelligence systems and, conversely, provides new computational frameworks for understanding biological processes. The bidirectional nature of this relationship suggests potential breakthroughs in both fields through cross-pollination of ideas and methodologies.
The synthesis of these concepts opens new avenues for research at the intersection of artificial intelligence, evolutionary biology, and developmental processes. This convergence not only enhances our understanding of natural evolutionary systems but also provides powerful new tools for developing more robust and adaptable artificial intelligence systems.

\begin{figure}
    \centering    
    \includegraphics[width=\linewidth]{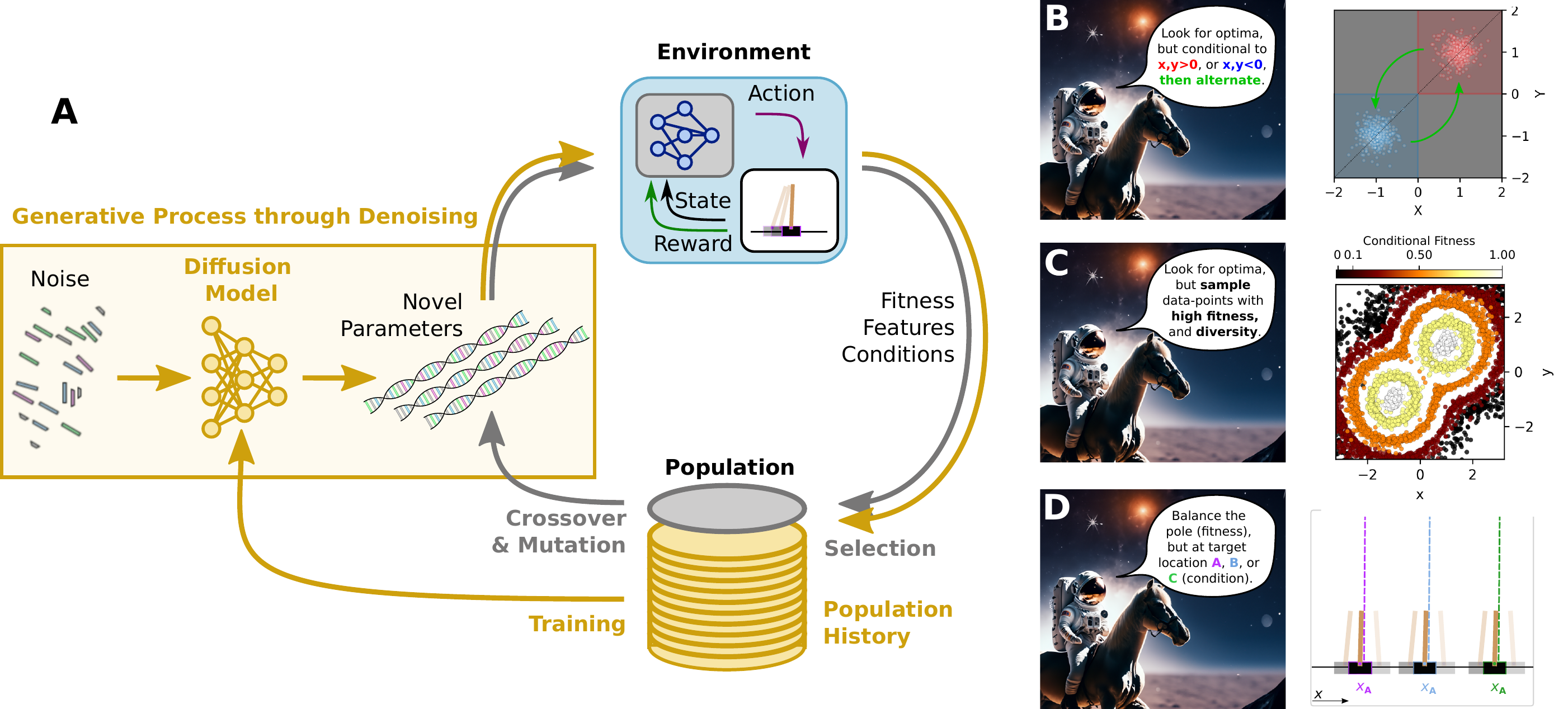}
    \caption{(\textbf{A}) A schematic flow-chart of a typical evolutionary algorithm (gray arrows and labels) contrasted with our diffusion model (DM)-based evolutionary optimization(golden arrows and symbols, see also \cref{algo:methods:charles}), showing an evolutionary process either utilizing population-based (gray) or an ANN-based DM (golden) as heuristically refined generative model for offspring-genotype sampling. The DM-based EA's generative model learns from heuristic experience by training on an epigenetic joint dataset of genome, fitness, and (potentially) conditional feature data of a particular genotype in its environment. We then utilize the successively refined DM to sample high-quality (high fitness) offspring candidate solutions for a particular environment; via classifier-free-guidance techniques~\citep{ho2022classifier}, this generative process can potentially be biased towards desired target traits in the environment on the phenotypic level.
    (\textbf{B}) Schematics of DM-based evolutionary optimization in an environment with two Gaussian optima at $\bm\mu_\pm=(\pm1,\pm1)$, but conditioning the search dynamics either to a target parameter range $x,y>0$ (red) or $x,y<0$ (blue), or alternate between the two peaks through dynamic conditioning  (green).
    (\textbf{C}) Schematics of utilizing conditional DM evolution of high-fitness genotypes (low to high fitness color-coded from black through orange to white) that maintains diversity (spread in parameter space).
    (\textbf{D}) Schematic behavior of DM-based conditionally evolved reinforcement learning (RL)~\citep{sutton1998reinforcement} agents deployed in the cart-pole environment~\citep{barto1983neuronlike}; the agents are evolved to maximize fitness (balance the pole), but conditionally sampled to steer the cart to a certain location (here A, B, or C) without changing the reward signal.
    }
    \label{fig:methods:CHARLES}
\end{figure}

Here, we empirically demonstrate that deep learning-based generative diffusion models can integrate genotypic reproduction processes from genetic algorithms to adapt within specific environmental and external conditions, guiding heuristic populations toward target parameter directions (see \cref{fig:methods:CHARLES}). More specifically, we utilize probabilistic diffusion models~\citep{ho2020denoising, song2020denoising}, based on Artificial Neural Networks (ANNs), to incorporate generational reproduction in evolutionary optimization (refer to \cref{fig:methods:CHARLES}~(A)). This approach allows us to explore diverse solutions in complex parameter spaces, achieving significantly higher evolutionary efficiency compared to conventional methods.

We emphasize, that, unlike previous work~\citep{yan2024EmoDM, krishnamoorthy2023diffusionmodelsblackboxoptimization}, we do not, in any way, pretrain the diffusion model on existing data, but constantly refine the generative process of the diffusion model on successively evidence acquired throughout the underlying evolutionary process.

Our work redefines evolutionary algorithms as a reproduction mechanism enhanced by diffusion models. We incorporate epigenetic associative memory into the evolutionary search process by iteratively training the diffusion model on a heuristically acquired dataset buffer after each generation. This associative memory promotes modular adaptations during reproduction, by capitalizing on prior experiences while minimizing inductive biases.

Additionally, through classifier-free guidance techniques~\citep{ho2022classifier}, we train generative diffusion models to conditionally associate high-quality parameters with specific genotypic, phenotypic, or even population-wide traits, independently of fitness scores (illustrated in \cref{fig:methods:CHARLES}~(B-D)). This strategy enables us to control the evolutionary process's search dynamics without altering the objective function but by conditioning the diffusion model to generate offspring with designated target traits or qualities.

Consequently, our approach supports biologically inspired multi-objective optimization without the need for complex reward-shaping techniques~\citep{Andrew1999PolicyInvarianceRewardTrafo}. This work establishes intriguing connections between conditional generative deep learning and evolutionary biology, demonstrating how memory and genotypic conditioning can influence evolutionary optimization algorithms to evolve lineages with specific genotypic, phenotypic, or population-wide characteristics.

Our findings suggest that the denoising process inherent in diffusion models can serve not only as a powerful tool for evolutionary optimization but also as a bridge between generative AI, diffusion models, and evolutionary programming—fields that may collectively advance toward more biologically plausible AI frameworks. 


\section{Methods}
\label{sec:methods}

\subsection{Evolutionary algorithms: Black-box heuristic optimization techniques}
\label{sub:methods:EA}
The principles of evolution have applications beyond biology, proving useful in addressing complex systems across different domains. The main components of this process - imperfect replication with heredity and fitness-based selection - are versatile and can be applied in diverse fields. In computer and data science, numerous optimization methods are employed; among the most widely used is stochastic gradient descent (SGD), which, with advancements like the Adam optimizer, excels in tasks where gradient calculations provide clear direction toward solutions. However, not all tasks crated equal, some are amenable to gradient-based methods, as gradient calculation can be intractable for many complex problems.  For these cases, Evolutionary Algorithms (EAs) such as CMA-ES~\citep{hansen2001completely} and PEPG~\citep{sehnke2010parameter} are essential. These heuristic optimization techniques~\citep{vikhar2016evolutionary, grefenstette1993genetic, goldberg1989genetic, holland1992adaptation} maintain and evolve a population of genotypic parameters over successive generations using biologically inspired operations, including selection, reproduction, crossover, and mutation. The goal is to gradually adapt the genotypic parameters of the entire population so individual phenotypic samples perform well when evaluated against an objective- or fitness function. The evaluated numerical fitness score of an individual correlates with its probability of survival and reproduction to drive the evolutionary process toward more optimal solutions. Thus, these algorithms use evolutionary biology dynamics to discover optimal or near-optimal solutions within vast, complex, and otherwise intractable parameter spaces. Such approaches are particularly valuable when heuristic solutions are needed to explore extensive combinatorial and permutation landscapes.

EAs can work with either discrete or continuous sets of parameters, with the former being a subset of the latter. Our focus here is on continuous parameter spaces that have domain-specific structures, which are typically \textit{a priori} unknown. As a result, the initial population is often sampled from a standard normal distribution. This population is then progressively refined with each generation to excel on a specific objective function. Essentially, the initially random parameters are heuristically adjusted by evolutionary algorithms, gradually transforming into highly structured parameters that perform effectively on the given task, with the goal of optimizing the objective function to solve the problem at hand.

The reproduction process of EAs to generate novel offspring parameters can either be population based through recombination and mutation operations at the genotypic level, or even leveraged by sampling novel data-points from successively re-parameterized probabilistic models, \eg, with a Gaussian prior~\citep{hansen2001completely}.
In essence, evolutionary processes thus act like generative models that are parameterized, or trained, based on heuristic information gathered from previously explored areas of the parameter space, at least from the prior generation and considering the current state of their underlying generative model. This setup is aimed at generating offspring that may be better adapted for the next generation. Furthermore, as we demonstrate in a complementary contribution~\citenum{zhang2024diffevo}, these evolutionary processes have similarities to diffusion models

\subsection{Integrating Diffusion Models as Offspring-Generative Process in Evolutionary Algorithms}
\label{sub:methods:DM}

The evolutionary process can be viewed as a transformation of genotype or phenotype distributions -- or, more generally, parameter distributions. The current population's distribution undergoes selection and mutation, evolving into a slightly different distribution after each time step. This perspective highlights the great potential of using generative models to simulate evolutionary dynamics. In particular, diffusion models, which have achieved state-of-the-art performance across various generative tasks--including image, video, and audio generation--are well-suited for modeling the complex distributional changes inherent in evolution. Given their capability to capture intricate data distributions, we hypothesize that a well-trained diffusion model should outperform traditional methods in evolutionary strategies.

Therefore, we propose diffusion model-based evolutionary strategies to generate offspring efficiently by sampling more from high-fitness regions in the parameter space. Given a population $\{\genome_{\gt,i}\}$ at time $\gt$, we evaluate their fitness values $f_{\gt,i} = f(\genome_{\gt,i})$ using a fitness function $f: \mathbb{R}^n \to \mathbb{R}$. Our key idea is to train a generative model on the current population and their associated fitness values, then sample the next generation $\{\genome_{\gt,i}\}$, which contains more high-fitness individuals. To achieve this, we map the samples $\{\genome_{\gt,i}\}$ and their fitness values $f_{\gt,i}$ to a density distribution and train a diffusion model on it. Specifically, we use a function $h: \mathbb{R} \to \mathbb{R}^+$ to map fitness to a fitness-derived probability density, such that $p(\genome) \propto h[f(\genome)]$.

Diffusion models consist of two phases: a forward diffusion phase and a reverse denoising phase. In the forward diffusion phase, noise is gradually blended into the training data--a process known as diffusion. A neural network is then trained to predict the added noise given the noisy data. In the reverse phase, starting from noisy data, the trained neural network is used to denoise step by step, eventually restoring the noise-free data.

Formally, during the forward diffusion phase, the noise-free data sampled from $\genome$ is considered at time zero, i.e., $x_0$. It is blended with noise over time according to:
\begin{equation}
    \vx_t = \sqrt{\alpha_t} \, \vx_0 + \sqrt{1 - \alpha_t} \, \ve,
    \label{eq:methods:dm:diffuse}
\end{equation}
where $\ve \sim \mathcal{N}(0, I^D)$, and $\alpha_t$ decreases monotonically from 1 to 0 as $t$ increases, with $\alpha_0 = 1$ and $\alpha_T = 0$. As a result, $\vx_T \sim \mathcal{N}(0, I^D)$ and $\vx_0 \sim \genome$.
More explicitly, we henceforth use the symbols $\genome$ and $\bm x_t$ to formally distinguish between genotypic parameters in the evolutionary process and parameters subjected to the diffusion model, respectively.
Thus, we consider noise-free (or denoised) data points $\bm x_0$ as genotypic parameters $\bm x_0\sim\genome$.

To denoise, a neural network $\epsilon_\theta$ is trained to predict the added noise by minimizing the prediction loss $L$:
\begin{equation}
    \theta = \min_\theta L(\theta) = \min_\theta \sum_{t=1}^T \sum_{\vx \in \mathbb R^n} p(\vx) \left\| \epsilon_\theta\left( \sqrt{\alpha_t} \, \vx + \sqrt{1 - \alpha_t} \, \ve, t \right) - \ve \right\|^2. \label{eq:explicit_form}
\end{equation}
Given the difficulty in obtaining the exact probability density $p(\bm x)$ for the training data, diffusion models are trained by sampling data points $\bm x \sim \genome$ and time steps $t \in [1, T]$. Therefore, the loss function can be reformulated without explicitly using $p(\bm x)$:
\begin{equation}
    L(\theta) = \mathbb{E}_{t \sim \mathcal{U}(0, T), \, \vx \sim \genome} \left\| \epsilon_\theta\left( \sqrt{\alpha_t} \, \vx + \sqrt{1 - \alpha_t} \, \ve, t \right) - \ve \right\|^2.
    \label{eq:methods:dm:loss}
\end{equation}
After training the model, the neural network $\epsilon_\theta$ can be used to sample new data points that follow the distribution of the training data. In the Diffusion Denoising Implicit Model (DDIM)~\citep{song2020denoising} framework, the sampling process is an iterative refinement:
\begin{equation}
    \vx_{t-1} = \sqrt{\alpha_{t-1}} \left( \frac{\vx_t - \sqrt{1 - \alpha_t} \, \ve_\theta(\vx_t, t)}{\sqrt{\alpha_t}} \right) + \sqrt{1 - \alpha_{t-1} - \sigma_t^2} \cdot \epsilon_\theta(\vx_t, t) + \sigma_t \, \ve_t,
\end{equation}
where $\ve_t \sim \mathcal{N}(0, I^D)$, and $\sigma_t$ is the noise amount. By default, we use $\sigma_t=\sqrt{(1-\alpha_{t-1})/(1-\alpha_t)}\sqrt{1-\alpha_t/\alpha_{t-1}}$. Starting with $\vx_T \sim \mathcal{N}(0, I^D)$, this iterative process generates $\vx_0 \sim \genome$.

To apply this to evolutionary tasks, we aim to assign higher probabilities to high-fitness individuals during sampling. Inspired by Equation~\eqref{eq:explicit_form}, we introduce a weighting function $h[f(\bm x)]$ into the loss function to bias the model towards high-fitness samples:
\begin{equation}
    L_{\text{evo}}(\theta) = \mathbb{E}_{t \sim \mathcal{U}(0, T), \, \vx \sim \genome} \, h[f(\vx)] \left\| \epsilon_\theta\left( \sqrt{\alpha_t} \, \vx + \sqrt{1 - \alpha_t} \, \ve, t \right) - \ve \right\|^2.
\end{equation}
With this modified loss function, the sampled data $x_0$ will follow a combined distribution of the original data and the fitness-derived distribution, i.e., $\vx_0 \sim h[f(\vx_0)] \, p_\genome(\vx_0)$, where $p_\genome$ represents the sample distribution.

We encapsulate this process into a function $\mathcal G$, which takes the current population $\genome_\gt$ and their fitness-derived density $p(\genome_\gt)$ to produce the new population $\genome_{\gt+1}$:
\begin{equation}
    \mathcal G: \left( \genome_{\gt}, \, p(\genome_{\gt}) \right) \to \genome_{\gt+1}.
\end{equation}
This mirrors the evolutionary process: at each step, we bias the sampling distribution $p_\genome$ with the fitness-derived probability density, effectively sampling more from high-fitness regions (selection). The inherent randomness in sampling introduces variation (mutation).

Different from our previous approach, by framing offspring generation as a generative process, we gain more control over this process. An important application is the use of conditional generation to steer evolution, including controlling population features, maintaining diversity, and even directly influencing fitness.

\subsection{Conditional, Heuristically-Adaptive ReguLarized Evolutionary Strategy through Diffusion (\CHARLES)}
\label{sub:methods:charles}

\begin{figure}
    \centering
    \includegraphics[width=1\linewidth]{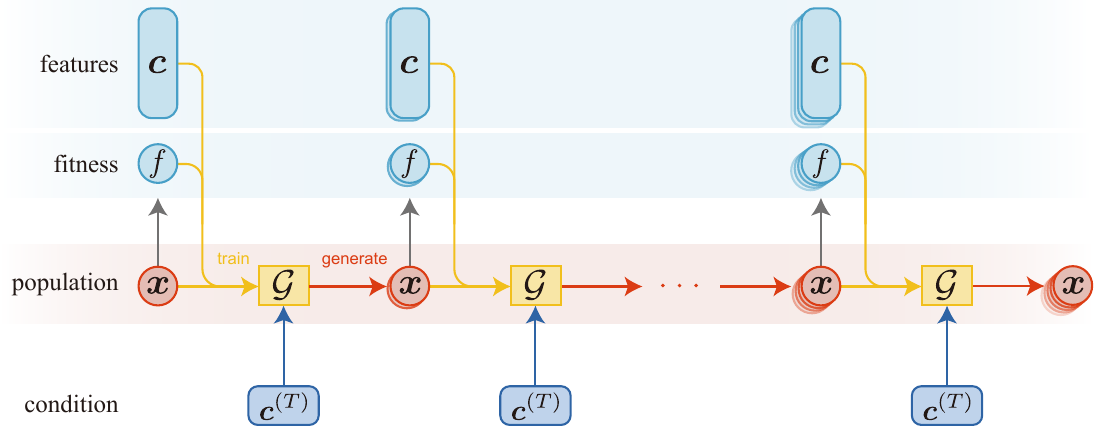}
    \caption{Workflow of the \CHARLES algorithm. Starting with a randomly initialized population (red circles), their fitness and features (shown in rounded blue rectangles) are evaluated. Next, a generative model $\mathcal G$ (yellow rectangles) is trained on this population, weighted by their fitness, with features used as conditioning for generation. Following training, externaltarget conditions are provided to generate a new population that meets specified requirements. The evaluation-training-generation loop is then repeated. A buffer is maintained to store the population along with their fitness and features for training, enabling full data utilization and preserving population diversity.
    }
    \label{fig:methods:CHARLES:details}
\end{figure}

DMs provide a model-free approach for learning denoising-based generative strategies tailored to custom datasets across versatile, problem-specific domains. Once trained on statistically relevant data, they can potentially surpass traditional EAs in generating high-quality offspring genotypes. In our complementary contribution~\citep{zhang2024diffevo}, we formally connect DMs to EAs and particularly demonstrate that the backward process in DMs can be viewed as an iterative evolutionary process across generations.

Here, we introduce a paradigm shift by sustaining and evolving a heuristic population  $\population_{\gt}=\{\genome_{\gt,1}, \genome_{\gt,2}, \dotsc, \genome_{\gt,\popsize}\}\sim p_{\dataset_{\gt-1}}$ that is sampled across successive generations $\gt=1,2,\dotsc,N_\gt$ from $p_{\dataset_{\gt-1}}$ via a heuristically refined diffusion model $\mathcal{G}_{\gt-1}\rightarrow\mathcal{G}_{\gt}$ (see \cref{sub:methods:DM}). This model is constantly refined - \ie, trained ``online'' - on a successively acquired dataset buffer containing elite solutions of previous generations $\dataset_{\gt-1}=\{\population_{\gt^\prime<\gt}\}$ that have been sampled by prior versions of the DM.
In training the DM, we notably weight high-fitness data more heavily compared to low-fitness genotypes using a fitness weighting function $h[f(\genome)]$. This approach increases the probability of sampling high-quality data while still maintaining diversity in the generative process (see \cref{sub:methods:DM,app:fitness:weighting:roulette} for details).
We then use $\mathcal{G}_{\gt+1}$ as a generative model for sampling high-quality genotypic parameters $\population_{\gt+1}\sim p_{\dataset_\gt}$ of the next generation $\gt+1$, with successively larger fitness than the prior generations.
We thus propose a ``Heuristically Adaptive Diffusion-Model Evolutionary Strategy'' (\HADES) for learning good and model-free reproductive strategies in EAs (compared to, \eg, the Gaussian prior in CMA-ES) by training DMs on fitness-weighted datasets of genotypic data; see pseudo-code in \cref{algo:methods:charles} and illustrations in \cref{fig:methods:CHARLES,fig:methods:CHARLES:details}.

This approach allows us to adaptively generate novel offspring parameters by refining either randomly initialized ``proto''-genotypes $\vx_T\sim\mathcal{N}(0,\sigmainit^D)$ (default), or recombined and mutated genetic material from elite solutions $i,j$ of the prior generation $\gt$, schematically expressed as $\genome_{\gt,i}\bigoplus\genome_{\gt,j}+\ve_t$ (see \cref{app:hades:sampling} for details).
Notably, the latter option of initiating the generative process of the DM with recombined genetic information relates to inpainting techniques~\citep{Lugmayr2022RePaint}, which aim to complete missing information in sample data. In traditional inpainting applications, missing information in masked or patched images is integrated seamlessly into the scene. Similarly, in our context, potentially conflicting parameter combinations from the genetic crossover operation (with potential effects on corresponding fitness scores) can be resolved by the DM, functioning in a manner akin to error-correction mechanisms~\citep{Frank2019}.

\begin{algorithm}
\caption{Pseudo-code of \HADES (\ccharles{\CHARLES}): novel generations are consecutively sampled by a heuristically refined DM. The seeds for this sampling are yet again sampled from crossover and mutation operations of the current population and from Gaussian noise to allow both adaptation of (combinations of) elite solutions and unbiased sampling by the DM. \tcharles{\CHARLES can be conditionally biased during sampling.} Notably, replacing the DM, $\mathcal{G}$, with a multivariate Gaussian model, \ccharles{and refraining from the option of using conditional sampling}, would essentially recover the CMA-ES~\citep{hansen2001completely} algorithm.}
\label{algo:methods:charles}
\begin{algorithmic}[1]
    \Require Population size $\popsize$, 
             parameter dimension $D$, 
             initial STD $\sigmainit$,
             fitness function $f$, 
             weighting function $h$, 
             diffusion model $\mathcal{G}$,
             crossover ratio $N_c$,
             total evolution steps $N_g$,
             \ccharles{classifier function $c$},
             \tcharles{target condition $\condition^{(\target)}$}.
    \Ensure $N_c<\popsize$
    
    \State $\population_1 \gets \mathcal N(0,\sigmainit^{\popsize{}\times D})$ \Comment{Initialize population}
    
    \State $\dataset \gets \{\varnothing\}$ \Comment{Initialize dataset buffer}
    
    \For{$\gt \in [1, 2, ..., N_\gt]$}
        \State $\{\genome_1, \genome_2, ..., \genome_{\popsize}\} \gets \population_\gt$
        \State $\forall i\in [1,\popsize]: 
               f_i\gets h[f(\genome_i; \gt)]$ 
        \State \ccharles{$\forall i\in [1,\popsize]: 
               \condition_i \gets c(\genome_i; \gt)$} 
        \State $\dataset \gets \dataset \bigoplus \{(\genome_i, f_i, \ccharles{\condition_i})\}$
               \Comment{Cache associated (data-point, fitness weight, \ccharles{and classifier})-tuples}
        \State $\mathcal{G}_{\gt} \gets \mathrm{train}(\mathcal{G}, \dataset)$
               \Comment{(Re)train diffusion model on updated dataset buffer}
        \State $\tilde{\population}_{\gt+1} \gets \mathrm{crossover}(\population_{\gt} | f_i, f_j)^{N_c} \bigoplus \mathcal{N}(0,\sigmainit^{(\popsize{}-N_c)\times D})$
               \Comment{Get crossover and noisy ``proto''-genomes}
        \State $\population_{\gt+1} \gets \mathcal{G}_{\gt}: p_\dataset(\tilde{\population}_{\gt+1} \tcharles{| \condition^{(\target)}})$
               \Comment{Sample next generation \ccharles{conditional to \tcharles{target} traits} via refined denoising}
    \EndFor
\end{algorithmic}
\end{algorithm}

Intriguingly, with DMs, we can apply techniques such as classifier-free guidance~\citep{ho2022classifier} to condition the generation process. This allows us to implement an evolutionary optimizer whose search dynamics can be controlled without relying on additional reward-shaping techniques~\citep{Andrew1999PolicyInvarianceRewardTrafo}. By training the DM with additional information $\condition_i = c(\genome_i)$, which numerically quantifies certain qualities or traits of genotypes $\genome_i$ in their respective environments, the DM learns to associate~\citep{Ambrogioni2023} elements in the parameter space with corresponding classifiers $\genome_i \leftrightarrow \condition_i$.

Technically, this is achieved by extending the input of the DM's ANN as $\epsilon_\theta(\vx_t, t) \rightarrow \epsilon_\theta(\vx_t, t, c(\vx_t))$. The function $c(\cdot)$ is a custom, not necessarily differentiable, vector-valued classifier function or a measurement of a trait of the data point $x$, or genotype $\genome$, evaluated in the parameter space, fitness space, or even phenotype space.

During sampling, the DM's generative process can be biased towards novel high-quality data points that exhibit a particular target trait $\condition^{(\target)}$, by conditioning the iterative denoising process of the diffusion model as $\epsilon_\theta(\vx_t, t, \condition^{(\target)}) \rightarrow \hat{\vx}_0$ such that $c(\hat{\vx}_0) \approx \condition^{(\target)}$. This allows the DM to generate high-quality samples with the desired traits, similar to how Stable Diffusion~\citep{rombach2022high} and Sora~\citep{brooks2024video} generate realistic image or video content based on custom text prompts.

In our context, we propose using conditional sampling to gain exceptional control over a heuristic search process with an open, successively refining dataset. While the heuristic nature of the evolutionary process facilitates global optimum exploration, the successively refined DM-based generative process allows for diverse sampling of high-quality genotypic data points that may exhibit target traits defined independently from the fitness score, akin to prompting an image-generative DM with text input. 

We consider this approach a ``Talk to your Optimizer'' application~\citep{Mathews2018} and refer to this method as \textit{Conditional, Heuristically-Adaptive Regularized Evolutionary Strategy through Diffusion} (\CHARLES). The pseudocode and a visualization are provided in \cref{algo:methods:charles,fig:methods:CHARLES:details}, where deviations of \CHARLES from the \HADES method in the \ccharles{training} and \tcharles{sampling} steps are color-coded for clarity.


\section{Results}
\label{sec:results}

\subsection{Diffusion models can efficiently learn, generate, and adapt genotypic representations within heuristic evolutionary processes}
\label{sub:results:competetive}
In the following section, we demonstrate that probabilistic diffusion models trained ``online'' on generations of heuristically varied parameter sets can be efficiently utilized as generative models in EAs, all based on successive denoising operations.
We first apply our technique to toy problems, specifically solving the dynamic and the static cases of the \textit{double peak} in \cref{sub:results:adaptive,sub:results:conditional:genotype,sub:results:conditional:dynamic} and \textit{Rastrigin} problem in \cref{sub:results:conditional:fitness,sub:results:conditional:novelty}, then we focus on Reinforcement Learning (RL) tasks~\citep{sutton1998reinforcement} in \cref{sub:results:conditional:phenotype}. 
We contrast the performance with other EAs, demonstrating that our method is competitive. 
Our approach enables the simultaneous identification of multiple optima, emphasizing diversity and exploration, while still furthering optimal solutions with high efficiency.
Furthermore, we demonstrate in the following that classifier-free guidance techniques~\citep{ho2022classifier} can be effectively utilized to conditionally bias the generative process in EAs. This approach constrains the search dynamics of the underlying heuristic evolutionary processes, even within complex rugged fitness landscapes, without changing the fitness score function. The method is analogous to the use of Lagrange multipliers in differentiable optimization, serving to impose constraints that effectively guide the search even in rugged, non-differentiable fitness landscapes.

In all our investigations, we use simple feed-forward neural network architectures for our diffusion models, which typically consist of \params{$2$} hidden layers with a number of hidden units ranging from \params{$24$-$324$}. To improve readability, we list all simulation parameters explicitly in \cref{app:simulation:details}.

\subsection{Diffusion models provide a model-free approach that adapts readily to new problems and dynamic environments}
\label{sub:results:adaptive}
\begin{figure}
    \centering
    \includegraphics[width=0.7\linewidth]{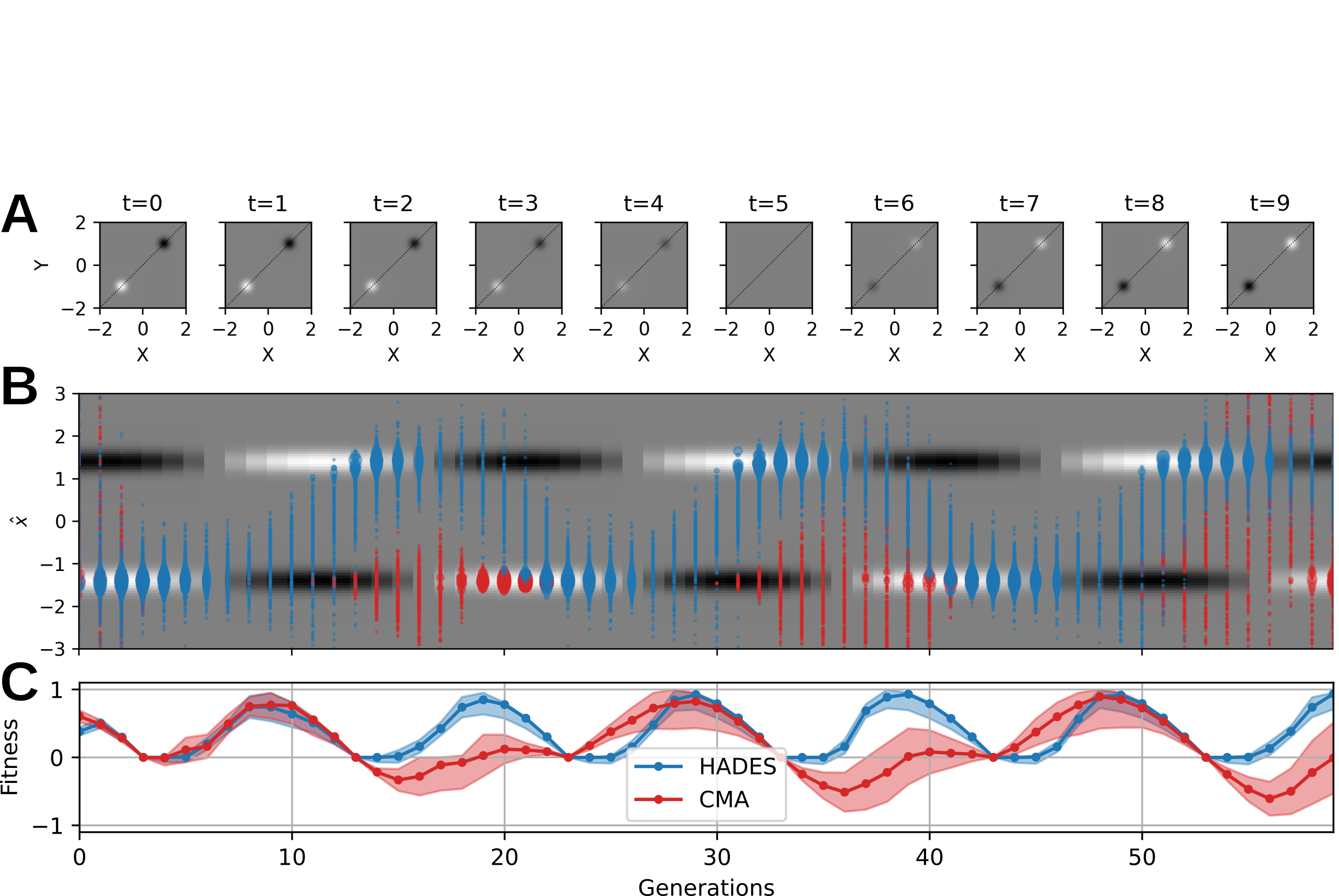}
    \caption{\HADES adapts to dynamic (oscillatory) environmental changes.
    (\textbf{A}) Dynamically alternating double-peak fitness landscape ranging from $f_{\min}=-1$ (black) through $f_\mathrm{0}=0]$ to $f_{\max}=1$ (white) as defined by \cref{eq:doublepeak:dynamic}.
    (\textbf{B}) Population data for \HADES (blue) and CMA-ES~\citep{hansen2001completely} (red) optimization in the dynamically changing environment illustrated in (A). The 2D data points $\genome_i=(x_i, y_i)$ are represented as 1D projections $\hat{\genome}_i$ onto the diagonal illustrated as dashed lines in (A); the background color indicates the fitness score along $x=y$, and the radius of the data-points $\hat{\genome}_i$ scales with fitness $f_i$, respectively.
    (\textbf{c}) Fitness of the data shown in (B): the solid line illustrates the maximum fitness evaluation of the population averaged over \params{10} statistically independent simulations; the shaded area illustrates the average spread of the population's maximum fitness. 
    While \HADES reliably identifies the current maximum in the alternating double-peak environment, CMA-ES clearly struggles to adapt a population to the changing environment in time as the majority of the population resides in the vicinity of one peak.
    }
    \label{fig:results:dynamic:env}
\end{figure}

Biological evolution is renowned  for its adaptability capabilities in (slowly) changing environments~\citep{levin2023darwin, Wagner2007, Schlosser2004, Calabretta2003, Tusscher2011, Wagner1996}.
While being increasingly recognized in the computational literature~\citep{power2015can}, mainstream EAs typically focus on exploring parameter spaces for globally optimal solutions to complex and often rugged, yet generally static objective functions.

To examine the  adaptive capabilities of evolutionary processes to changing environmental conditions to mimic a more biologically  realistic setting, we introduced a time-dependent objective function, $f(\genome,\gt)$, and compare the learning capabilities of \HADES against other different mainstream EAs.
Specifically, we define $f(\genome,\gt)$ as
\begin{equation}
    f(\genome,\gt)=\cos(\omega\,\gt)\,e^{-\frac{(\genome-\mathbf{\bm\mu_-})^2}{2\sigma^2}} + \cos(\omega\,\gt + \phi)\,e^{-\frac{(\genome-\bm{\mu_+})^2}{2\sigma^2}},
    \label{eq:doublepeak:dynamic}
\end{equation}
with two Gaussian peaks of STD $\sigma$ centered at $\bm\mu_{\pm}=(\pm1,\pm1)$ with amplitudes that oscillate phase-shifted by $\phi$ across generations $\gt$ with angular velocity $\omega$; a static double-peak problem is recovered by setting $\omega=0$ and $\phi=0$.

Thus, we utilize an alternating double-peak function, where one peak has a positive and the other one negative amplitude by setting \params{$\phi=\pi$, $\omega=2\pi/10$, and $\sigma=0.1$}. Over time, the amplitudes periodically alternate in sign, reverting the target of the maximization objective. 
We apply our \HADES method and CMA-ES~\citep{hansen2001completely} both with a population size of \params{$\popsize=256$} and an initial population of standard deviation (STD) \params{$\sigmainit=0.5$} evaluating each sampled individual $\genome_i$ in every generation $\gt$ against $f(\genome_i,\gt)$; see \cref{app:tab:simulation:details} in \cref{app:simulation:details} for more details.
The population dynamics are depicted in \cref{fig:results:dynamic:env}~(B).
Traditional EAs have very different strategies in updating their generative models across generations due to their inherent inductive biases. Yet, even powerful mainstream approaches such as CMA-ES may fall short in adapting a population to changing environmental conditions after having seemingly converged on a solutions, even for simple problems as the alternating double-peak function discussed here.
In contrast, \HADES consistently identifies the periodically changing maximal fitness peak by adapting its population by sampling new offspring through an ever-refined DM; this is reflected by the respective time-dependent fitness of both approaches depicted in \cref{fig:results:dynamic:env}~(C).

Thus, our \HADES method offers an efficient and model-free approach to generate high-quality genotypic data, particularly excelling in scenarios that require enhanced adaptability capabilities. 
The intrinsic representative power of DMs enables reliable learning of subtle signals and correlations within arbitrary parameter sets, while their versatile sampling capabilities during the generation phase allow for precise control over the output characteristics. This combination of robust correlation learning and a flexible generative process makes \HADES particularly well suited to explore complex genotypic landscapes where traditional approaches might struggle to maintain both diversity and quality.

\subsection{Insights from developmental biology: Neutral multi-objective adaptation via conditional diffusion model evolution}
\label{sub:results:conditional:genotype}
\FloatBarrier
\begin{figure}
    \centering
    \includegraphics[width=0.7\linewidth]{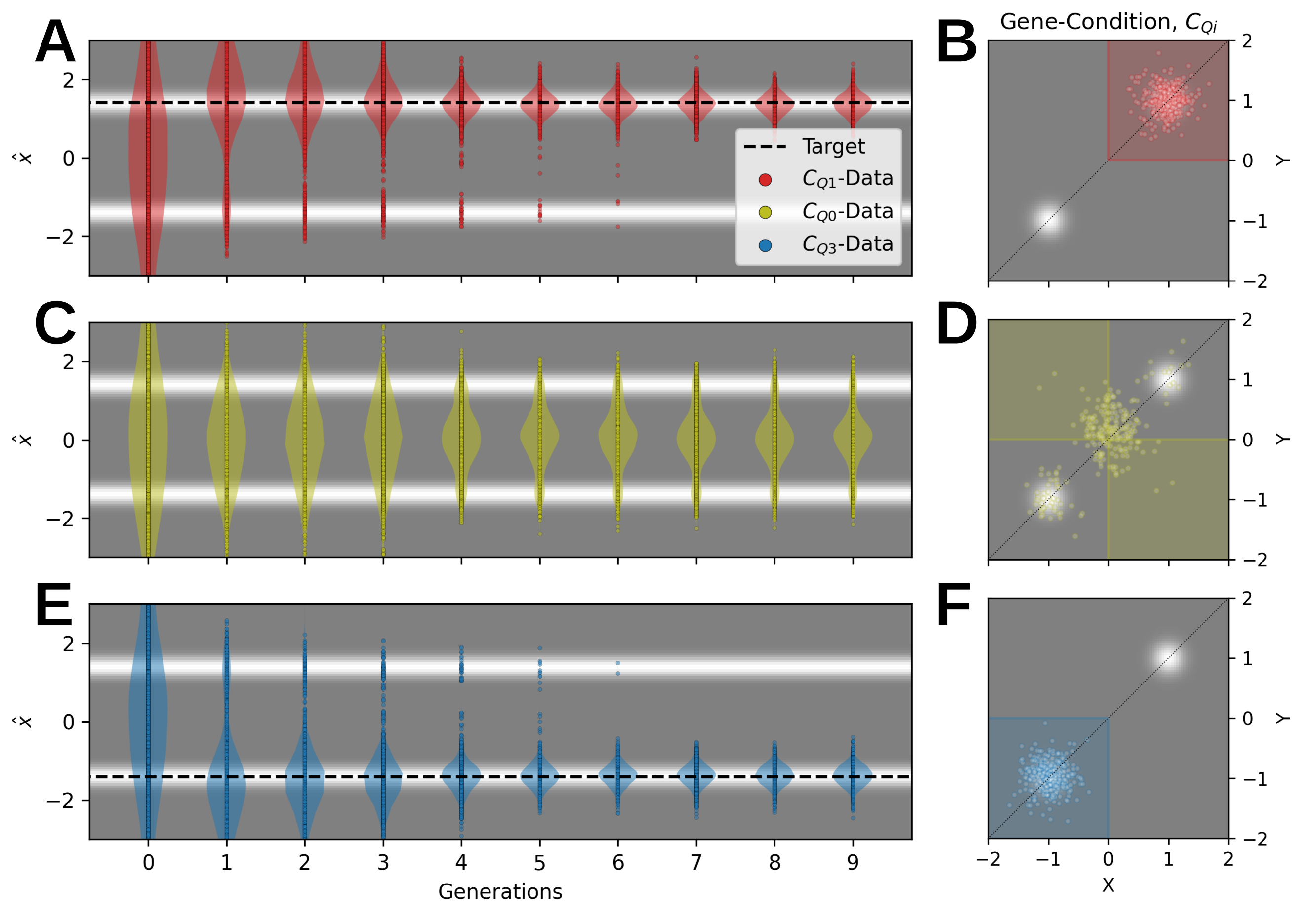}
    \caption{Conditional evolutionary optimization to explore selected target parameter regions in two-dimensional double-peak fitness landscape.
    (A, C, E) Fitness landscape (grayscale) and distribution of population data (projected onto the $x=y$ line) for \params{10} statistically independent simulations vs. generations as violin plot, while conditioning the generative DM to sample novel data points from the first quadrant $x,y>0$ (red), second and forth quadrants $x\times y <0$ (yellow), and third quadrant $x,y<0$ (blue); datapoints are projected onto the $x=y$ diagonal.
    (B, D, R) Fitness landscape in gray-scale from $f_{\min}=0$ (gray) to $f_{\max}=1$ (white) with overlaid data-points (colored dots) of an exemplary population from panels (A, C, E), respectively, after \params{9} generations; the target quadrants are illustrated as color-shaded areas.
    }
    \label{fig:results:conditional:genotype}
\end{figure}

Typically,  EAs aim to find optima in their respective fitness landscapes by maximizing fitness scores through biologically inspired selection and mutation operations.
In rugged fitness landscapes, this approach can be exceptionally effective, enabling exploration for global solutions, a distinct advantage over many gradient-based methods that often get stuck at local optima~\citep{Katoch2020, Sutskever2013Momentum}.
However, traditional EAs often struggle with problems that have multiple (unrelated or competing) objectives, as these can create conflicts and frustration in fitness scoring. While mitigation strategies such as problem-specific reward shaping~\citep{Andrew1999PolicyInvarianceRewardTrafo} and curriculum learning techniques~\citep{Bengio2009CurriculumLearning} exist, these approaches typically demand careful customization and domain expertise. This limitation highlights the need for more robust and adaptable optimization frameworks that can naturally handle multi-objective scenarios.

We draw inspiration from developmental biology to propose an alternative approach. Recent work~\citep{levin2023darwin}, suggests that biological evolution does more than just create organisms adapted to specific environments: it produces versatile problem-solving systems. These biological systems — ranging from gene networks to cells, tissues, organs, organisms, and even organismal collectives, demonstrate remarkable capacity to adapt to environmental cues or domain-specific challenges in real-time, while maintaining their overall physiological integrity~\citep{levin2022technological}.
Especially during an organism's development, but also during its life time, this adaptation  to environmental constraints manifests through sophisticated response mechanisms across multiple scales, while preserving core functionalities without compromising the organism's fundamental fitness. We can interpret this as a form of physiological conditioning of the multi-scale generative processes of biological systems allowing them to adjust to environmental constraints neutral to their system-level fitness. The universality of this principle suggests its applicability to mechanisms at the level of RNA and DNA.

DMs offer a particularly suitable framework for incorporating external (environmental) cues in their generative process through classifier-free guidance~\citep{ho2022classifier}:
At their most basic level, DMs are trained to generate, \ie, sample novel data $\hat{\vx}_0$ from a probability distribution $p(\vx)$ that conforms to a training dataset $\dataset=\{\genome_1, \dotsc, \genome_N\}$. The true power of DMs, however, lies in their ability to conditionally sample novel data points $\hat{\vx}_0^{(c)}$ that exhibit desired target traits or features $\condition$. 
These feature vectors $\condition$ numerically classify selected qualities of the data, which are here formally encoded via a custom, not necessarily differentiable mapping $\condition=c(\genome)$.
Thus, through classifier-free guidance, we can explicitly steer the generative process of DMs to sample biased data points $\hat{\vx}_0^{(c)}$ from a conditional probability distribution $\hat{\vx}_0^{(c)}\sim p(\vx|\condition)$, where the generated output exhibits specific desired target traits such that $\hat{\mathbf{c}}=c(\hat{\vx}^{(c)}_0)\approx \mathbf{c}$, see \cref{sub:methods:charles} for details.

This approach forms the foundation of modern text-guided image and video generation systems~\citep{brooks2024video, rombach2022high}. During training, the DM learns to associate~\citep{Ambrogioni2023} data $\genome$ with corresponding numerical conditions $\mathbf{c}=c(\genome)$. When deployed, it can generate data conforming 
to these learned conditions, effectively translating abstract constraints into concrete output characteristics.

Here, we employ classifier-free guidance to constrain the genotype sampling process in our \HADES method, ensuring that resulting phenotypes meet specific target conditions $\condition^{(\target)}$ in their environments. Crucially, this process operates independently of fitness scores!
We call this approach \textit{Conditional, Heuristically-Adaptive ReguLarized Evolutionary Strategy through Diffusion} (\CHARLES), which, in contrast to \HADES, modifies the denoising process by incorporating target conditions into the error estimate: $\epsilon_\theta(\vx_t, t, \condition^{(\target)})$.
The implementation involves three key steps:
First, we associate each element $\genome_i$ in the heuristic training data $\dataset$ with a numerical feature vector $\condition_i=c(\genome_i)$ that quantifies specific traits. These traits can encompass genotypic parameter qualities (see \cref{sub:results:conditional:genotype,sub:results:conditional:dynamic}), fitness-related metrics (\cref{sub:results:conditional:fitness}), population-level characteristics (\cref{sub:results:conditional:novelty}), or phenotypic traits (see \cref{sub:results:conditional:phenotype}).
Second, at each generation, we train the DM on paired data ${(\genome_i, \condition_i)}$; the loss function evaluates the model's ability to denoise data $\genome_i$ at fixated corresponding features $\condition_i$ that are unaffected by noise.
Third, during the generative phase of a new populations, we employ custom target conditions $\condition^{(\target)}$ to guide evolutionary trajectories through the parameter space. This leads to conditionally sampled individuals that increasingly exhibit the desired condition-specific traits while improving their fitness across generations.

In our first example, we apply the \CHARLES method to find optimal solutions to the static double-peak objective function given by \cref{eq:doublepeak:dynamic} with \params{$\omega=\phi=0$}, while conditioning the generative process of the DM to predominantly sample genotypic offspring in specific quadrants of the two-dimensional plane. The results are illustrated in \cref{fig:results:conditional:genotype}.
Starting from a normal distributed population \params{$\sigmainit=2$}, we associate individual genotypic parameters $\genome_i=(x_i, y_i)$ with their corresponding quadrant in the parameter space:
For data points in the first quadrant $(x_i,y_i>0)$, we assign the numerical classifier value $\condition^{(Q1)}_i=1$, for data points in the second and fourth quadrant $(x_j\times y_j<0)$, we assign $\condition^{(Q0)}_j=0$, and for data points in the third quadrant $(x_k,y_k<0)$, we assign $\condition^{(Q3)}_k=-1$.
The process then proceeds through three sequential steps:
First, we evaluate the fitness and feature vectors for all individuals of a given generation.
Second, the DM undergoes joint training on the set of associated data and quadrant-classifiers ${(\genome_i,\condition_i^{(Qj)})}$.
Finally, when sampling the genotypes of the next generation, we select a particular target quadrant, $T$, to condition the generative phase of the DM by $\condition^{(QT)}$.

In \cref{fig:results:conditional:genotype}~(A, C, E), we present population data across consecutive generations in this double-peak environment for \params{$10$} statistically independent evolutionary lineages. We condition the \CHARLES method to sample from one of three distinct regions: the first quadrant, $T=1$, the second and fourth quadrants, $T=0$, or the third quadrant, $T=3$. Snapshots of converged generations for each case are superimposed on the fitness landscape in \cref{fig:results:conditional:genotype}(B,D,F).
The results demonstrate the remarkable effectiveness of conditioning on the first and third quadrants:  despite both fitness peaks being the qualitatively equivalence, the \CHARLES method consistently converges to the peak located in the conditionally targeted quadrant. Conversely, conditioning DM-sampling on $\condition^{(Q0)}$ results in frustration effects, as the second and fourth quadrants lack fitness peaks.
We conclude that conditioning in \HADES serves as a powerful regularizer for genotypic exploration with DM evolution. This mechanism enables selectively biasing the evolutionary process to either explore or avoid specific regions of the parameter space, or more broadly, solutions with particular genotypic qualities, without altering the problem's fitness score. This neutral adaptation toward desired qualities is achieved elegantly through DM conditioning with classifier-free guidance techniques.
Furthermore, these conditions can be formulated flexibly and orthogonally to the optimization problem's fitness-score, without requiring differentiability. Consequently, \CHARLES provides an elegant approach to multi-objective optimization that eliminates the need for cumbersome reward shaping techniques~\citep{Andrew1999PolicyInvarianceRewardTrafo}.

\subsection{Learning from Past Experience: DM-Based Generative Samplers Provide a More Powerful and Biologically Inspired Framework for Modular Evolutionary Processes}
\label{sub:results:conditional:dynamic}
\FloatBarrier

\begin{figure}
    \centering
    \includegraphics[width=0.7\linewidth]{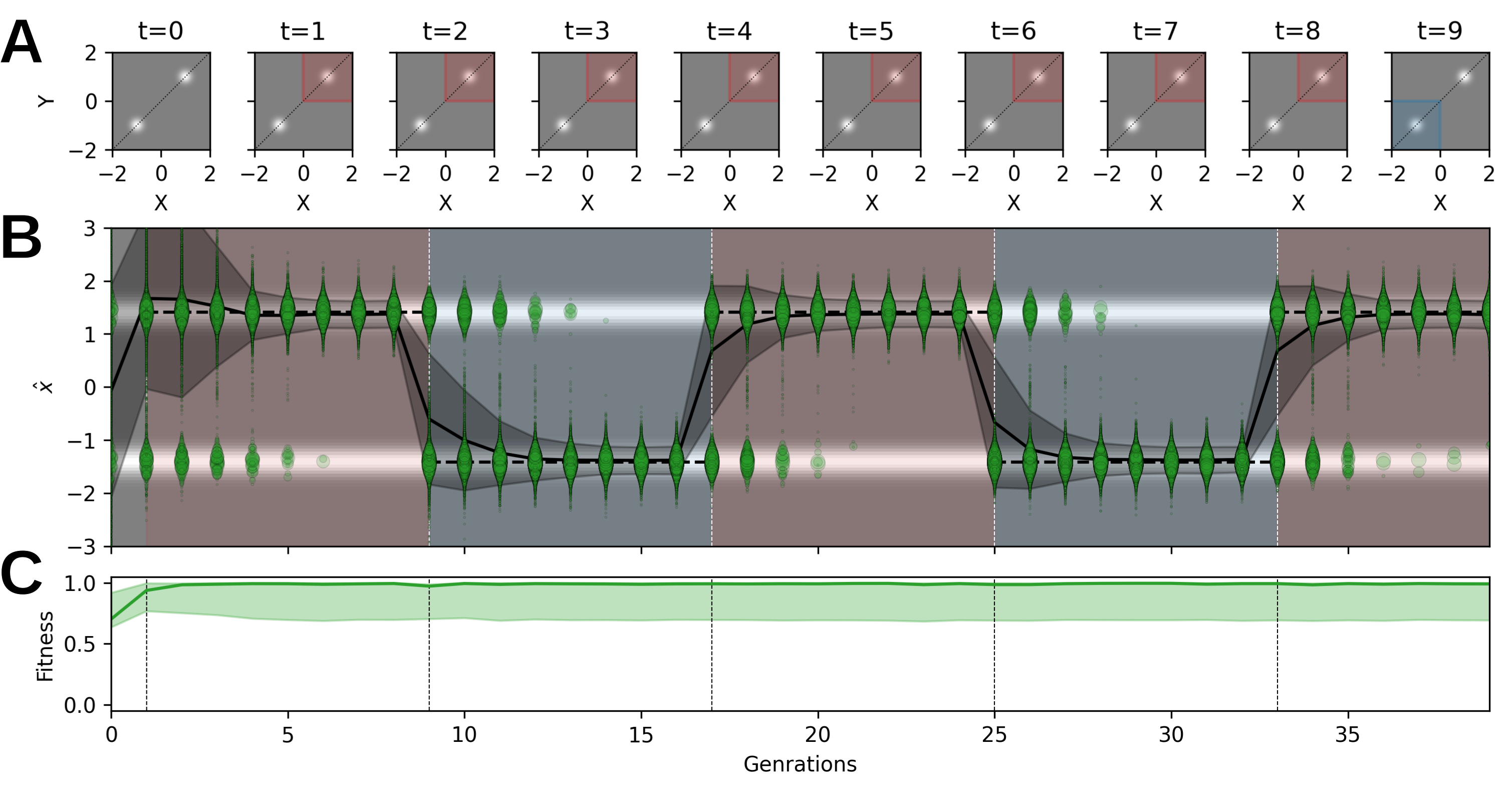}
    \caption{Dynamically Conditioning Genetic Parameters.
    (A) A static double-peak fitness landscape, ranging from $f_{\min} = 0$ (gray) to $f_{\max} = 1$ (white) as defined by \cref{eq:doublepeak:dynamic} with \params{$\omega=\phi=0$}. Dynamical conditioning allows exploration of the first quadrant (red) or the third quadrant (blue), see also \cref{fig:results:dynamic:env}.
    (B) The fitness landscape (grayscale) and the distribution of population data (projected onto the $x=y$ line) for \params{10} statistically independent simulations vs. generations (radii of green-colored data-points scale with fitness). 
        The average population mean is illustrated by the black solid line, while the gray area marks the STD.
        The oscillating red $\leftrightarrow$ blue color-coding of the fitness landscape reflects the applied condition for the first or third quadrant during DM sampling, respectively leading to jumps of the population from one peak to the other;
        transition generations are marked by white vertical dashed lines.
    (C) The mean and STD of maximum fitness (solid green line and shaded area) demonstrate consistently  high fitness values, even during transitions of the conditional sampling.
    }
    \label{fig:results:conditioning:dynamic}
\end{figure}

Biological evolution exhibits  an inherent capacity to utilize existing, heuristically acquired knowledge for species adaptation in changing environments~\citep{Frank2009a, kouvaris2017evolution, watson2016can, watson2016evolutionary, power2015can, Watson2014}. This capability stems from the recombination of established genetic material across evolutionary time-scales. Thus, the exploitation of previously explored solutions and their adaptation to novel contexts represents a fundamental principle of biological evolution~\citep{Wagner2007, Schlosser2004, Calabretta2003, Tusscher2011, Wagner1996}.
Within traditional EAs, the integration of past experience remains largely limited to heuristic information stored in the current population, or within the generative model employed by techniques such as CMA-ES for offspring sampling. Consequently, the generative process in current EAs maintains limited memory of previously explored solutions, leading to inefficiencies from either memory loss or the intrinsic inductive bias of generative models.
Generative DMs, however, can be equipped with \textit{epigenetic memory} by preserving information about previously explored solutions. This capability can be implemented through dataset buffering or DM retraining across multiple generations. These approaches offer a model-free methodology to enhance evolutionary sampling capabilities based on past experience, while simultaneously testing new environmental hypotheses through an iteratively updated generative process.
Information holds significant value in optimization, particularly in computationally expensive tasks where maintaining reliable memory is crucial. Refining a model-free generative process via heuristically acquired dataset buffers therefore presents an efficient approach to leverage prior knowledge in evolution. This significantly improves the capacity of an evolutionary process (as modeled by \HADES and \CHARLES) to navigate complex parameter spaces more effectively. Moreover, this ability to learn from past experience substantially augments the adaptability and transferability of evolutionary algorithms to changing environments.

To demonstrate this, we conduct the following experiment:
(i) utilize the static double-peak environment defined in \cref{sub:results:conditional:genotype},
(ii) equip \CHARLES with a memory buffer dataset spanning the past \params{$5$} generations, where only the lowest fitness solutions are replaced with current population data,
and (iii) implement time-dependent conditioning that alternates between the first and third quadrants in the parameter space.
We again start with an initial normal distribution population with \params{$\sigmainit=1$}. We then conditionally sample solutions from the first quadrant for \params{eight} consecutive generations using \params{$\condition^{(QT)}=\condition^{(Q1)}$ for $\gt\in[1,8]$} during the DM's generative process, followed by conditioning to the third quadrant for the next \params{eight} consecutive generations using \params{$\condition^{(QT)}=\condition^{(Q3)}$ during $\gt\in[9,16]$}, as shown in \cref{fig:results:conditioning:dynamic}(A). This process repeats periodically for each lineage, with DM initialization and training on the memory buffer dataset after each generation. The experiment is repeated \params{$10$} times, with population dynamics and elite fitness-scores presented in \cref{fig:results:conditioning:dynamic}(B,C).
The results reveal that \CHARLES consistently converges to the specified target peak, maintaining performance even under dynamically modified conditions during evolution. Notably, the epigenetic memory enables the evolutionary process to perform discontinuous but targeted transitions between previously visited high-fitness regions when conditions switch between $\condition^{(Q1)}\leftrightarrow \condition^{(Q3)}$. Rather than exhibiting slow continuous adaptation between peaks located at $\pm\bm\mu$, we observe instantaneous repopulation of the new target peak and heuristic fade-out of the previous one. This rapid, targeted readaptation maintains consistently high fitness scores across conditioning transitions (see \cref{fig:results:conditioning:dynamic}(C)), contrasting with the oscillatory elite fitness scores observed in the dynamically changing environment discussed in \cref{fig:results:dynamic:env}(C).
This behavior bears striking resemblance to the ecological memory described in Ref.~\citep{power2015can}, though achieved here through a heuristically adaptive DM.

In conclusion, our experiments demonstrate the powerful capability of generative DMs to leverage  evolutionary history for enhanced optimization processes. The integration of memory components and conditioning schemes in \CHARLES and \HADES significantly improves the adaptive capabilities of evolutionary algorithms in dynamic environments.
Our findings extend beyond algorithmic improvements: they raise fundamental questions about unconventional memory mechanisms in biological systems. The observed parallels between DM-based evolutionary processes and biological adaptation mechanisms suggest promising new directions for understanding the principles governing both artificial and natural evolutionary systems.

\subsection{Improving on Past Experience: Conditioning for Higher Fitness Can Improve Learning Performance, but Increases Greediness}
\label{sub:results:conditional:fitness}

Next, we investigate the application of fitness conditioning during the DM's generative process: We can jointly train the DM on associated parameters and fitness scores $\{\genome_i, f_i=f(\genome_i)\}$, conditioning the sampling process to generate offspring that potentially achieve a higher target fitness than any previously observed~\citep{krishnamoorthy2023diffusionmodelsblackboxoptimization}, $f^{(T)}>\max_i{(f_i)}$.

Given that the maximum fitness or reward can not be known \textit{a priori}, we propose sampling the target fitness for conditionally generating the next generation based on Fishers fundamental theorem of natural selection, which states that~\citep{Fisher1930} ``The rate of increase in fitness of any organism at any time is equal to its genetic variance in fitness at that time''.
This principle on the rate of expected fitness improvement has intriguing links to inverse reinforcement learning (IRL)~\citep{schmidhuber2020reinforcementlearningupsidedown, srivastava2021trainingagentsusingupsidedown}, as it can not be assumed that conditioning the DM on arbitrarily large fitness (\ie, significantly surpassing the training data) will yield reasonable offspring parameters, especially in early stages of the evolutionary search process.

However, to avoid this algorithm to become too greedy, we here introduce two flavors of fitness sampling, 
(i) Fisher-conditioning $\condition^{(TF)}\sim \mu_f + |\mathcal{N}(\mu=0, \sigma=\sigma_f)|$, 
and (ii) Greedy-conditioning of target fitness $\condition^{(TG)}\sim \mathcal{N}(\mu=f_{\max}, \sigma=\sigma_f)$, where $\mu_f$ is the mean, $f_c$ the maximum, and $\sigma_f$ the STD of the fitness scores $f_i$ of the current population. 

Illustrative examples of sampling parameters conditional to target fitness values by a DM pre-trained~\citep{krishnamoorthy2023diffusionmodelsblackboxoptimization} on the double-peak task are shown in \cref{fig:methods:CHARLES}~(C).
In \cref{fig:results:conditioning:novelty} (A-D), we present optimization results for different configurations of \HADES and fitness-conditional \CHARLES applied to the inverted and truncated Rastrigin task (see \cref{app:fitness-landscapes:rastrigin}), a periodically oscillating function in the two-dimensional plane with four optima located at $|x|=|y|=3.5$. For simplicity, we henceforth refer to the inverted and truncated Rastrigin task simply as Rastrigin task.

Our findings indicate that this fitness conditioning can indeed significantly improve the learning capabilities of \CHARLES, which opens new avenues for exploring complex parameter spaces.
However, aiming for greater fitness increases across generations might also lead to more greedy behavior of our algorithm and thus suboptimal convergence, which we will delve into detail in the following section.

\subsection{Novelty-Conditional Sampling: A Good Regularizer for Exploration and Maintaining Diversity}
\label{sub:results:conditional:novelty}
Typically, EAs aim at identifying solutions with optimal fitness values in rugged parameter landscapes. However, depending on the fitness landscape, this can be highly non-trivial and requires a dedicated balance between exploration and exploitation, see for example~\citep{HazanLevin}.
Recent discussions, originating from developmental biology~\citep{levin2023darwin,levin2022technological} suggest that biological systems operate under a different paradigm, where agents continuously explore novel situations to maintain their integrity and adapt to changing environments. This inherent drive for novelty leads to the creation of novel challenges, requiring further adaptation, and thus creating a diversification scaffold~\citep{Levin2024SelfImprovisingMemory}. 

\begin{figure}
    \centering
    \includegraphics[width=0.49\linewidth]{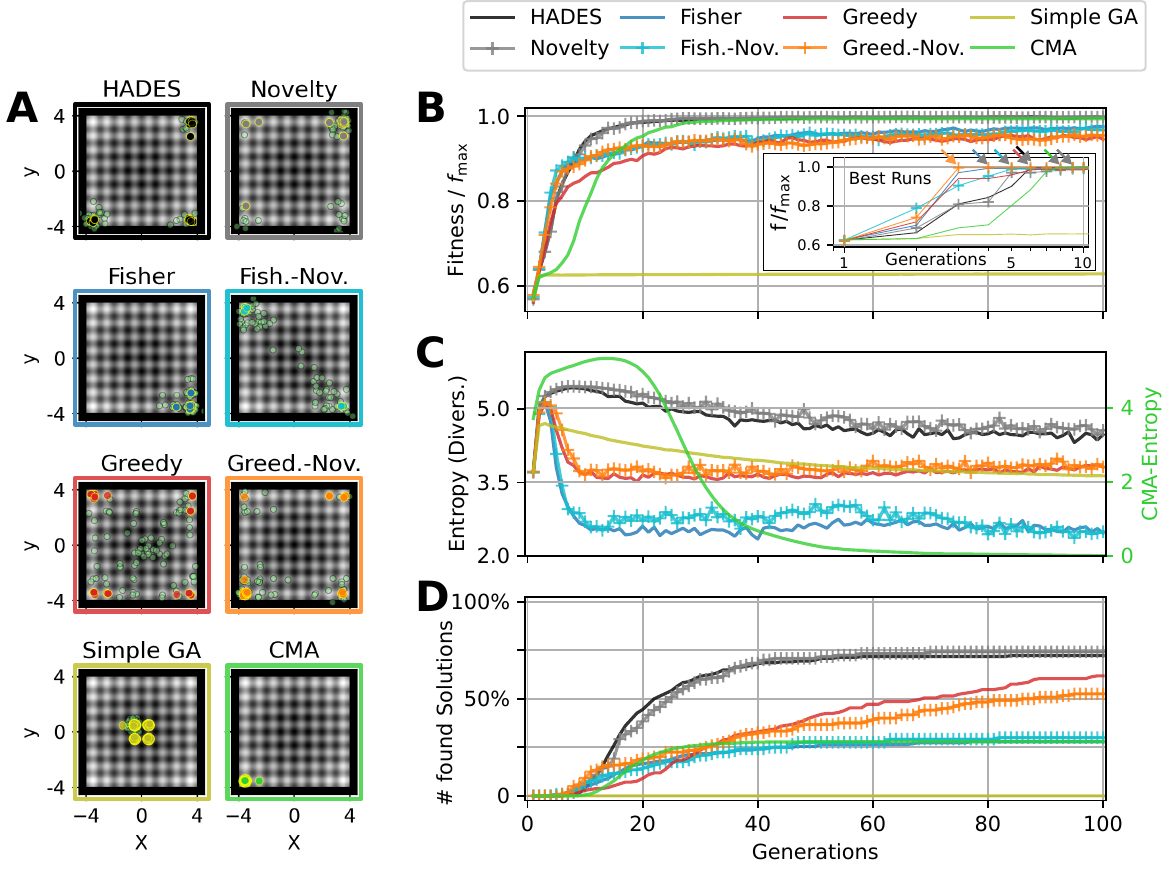}
    \includegraphics[width=0.49\linewidth]{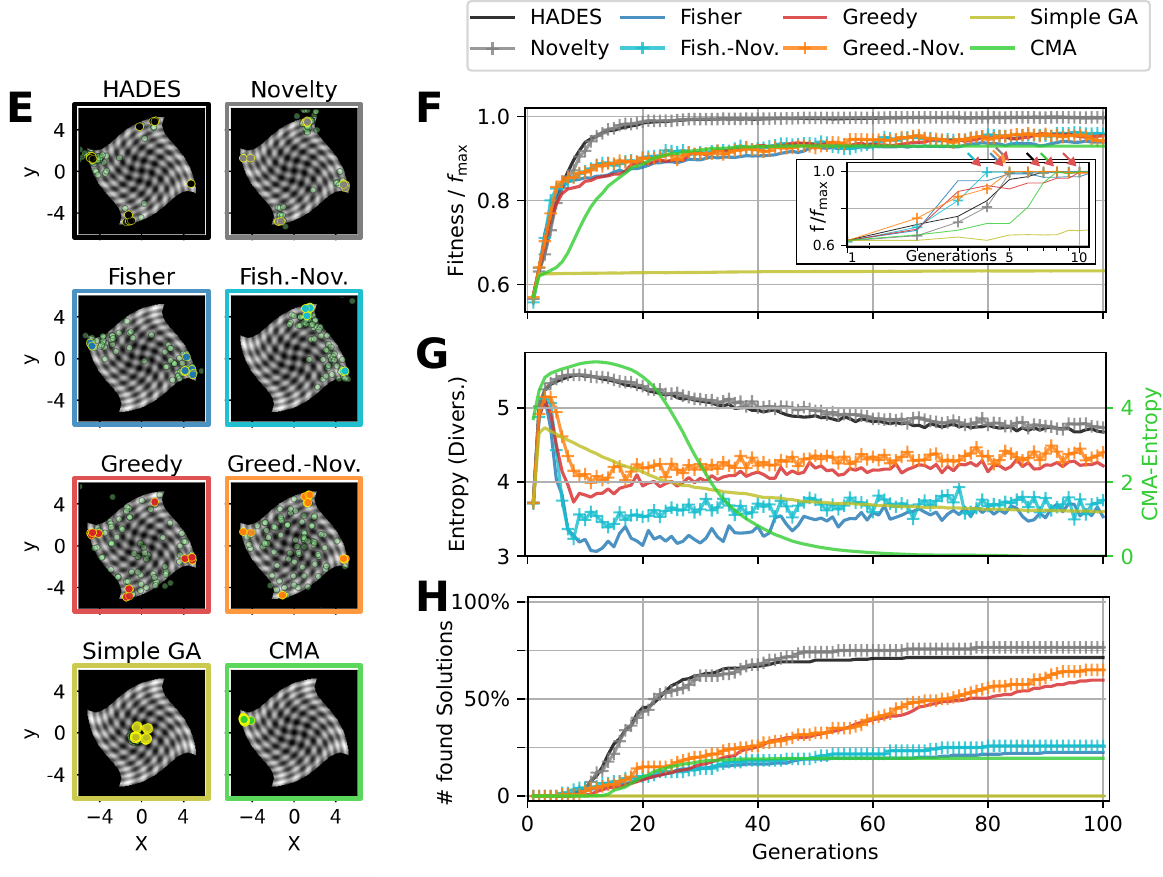}
    \caption{Fitness- and Novelty-Conditional Benchmarks for the Rastrigin (\textbf{A-D}) and ``Twisted''-Rastrigin (\textbf{E-H}) Task (see \cref{app:fitness-landscapes:rastrigin} for details).
    (\textbf{A}, \textbf{E}) The fitness landscape of the Rastrigin (A) and ``twisted''-Rastrigin task (E) with overlaid exemplary results (elite solutions in thick colored-coded circles, and final population in small green circles) after \params{$100$} generations of optimization with different \HADES and \CHARLES solvers (see titles and colored-coded borders) contrasted with SimpleGA and CMA-ES algorithms (bottom row, see text); the fitness landscape is indicated gray scale ranging from $f_\mathrm{min}=0$ (black) to \params{$f_\mathrm{max}=64.62$} (white). 
    (\textbf{B}, \textbf{F}) The maximum fitness from different solver configurations (\cf, color-coding and panels (A) and (E), respectively), 
    (\textbf{C}, \textbf{G}) the entropy-based diversity of the population (see \cref{app:entropy:estimate}), and 
    (\textbf{D}, \textbf{H}) the number of cumulatively identified solutions across successive generations for the Rastrigin (B-D) and twisted-Rastrigin (F-H) tasks, averaged  over \params{50} statistically independent simulations, respectively.
    The instets in (A, E) show the cumulatively best fitness vs. generations from all statistically independent simulations for the different solvers, and the colored arrows indicate the generation when the problem was solved in the least number of generations by a particular solver for both tasks.
    Both the vanilla \HADES and the novelty-conditional \CHARLES methods reliably identify multiple optima at the corners of the fitness landscape (D, E): on average, \params{$\approx75\%$} of the target peaks are identified, while in \params{$\approx 10\%$} of all simulations $100\%$ of the solutions are found successfully. The diversity measures shown in (C, G) and (D, H) indicate that the novelty-conditional \CHARLES method maintains an increasingly diverse populations of high-quality genetic material compared to the respective unconditional cases.
    }
    \label{fig:results:conditioning:novelty}
\end{figure}

It has recently been argued~\citep{Levin2024SelfImprovisingMemory}, that the fundamental drive for novelty-biological systems could be rooted in the phenomenon of boredom: biological agents on a variety of scales will seek novel stimuli if their sensory inputs stagnate too long. This is reflected in their intrinsic drive for exploration and discovery, but not necessarily aligns with traditional optimization objectives or search strategies. 
In fact, incorporating novelty~\citep{lehman2011abandoning} or a bias toward quality-diversity~\citep{pugh2016qualitydiversity} into optimization processes such as EAs significantly improve their performance. 
These techniques reinforce search directions in novel parameter regions while penalizing (over)exploitation of already experienced solutions. 
Instead of directly optimizing for novel and diverse solutions via modified fitness scores, we propose to use novelty-conditional sampling via \HADES.

We utilize a metric for diversity in the heuristic data buffer of our \CHARLES method, and condition the DMs generative process to sample diverse genotypes.
Specifically, we define diversity $\delta_i$ of a single data point $\genome_i$ similar to the non-parametric $k$-nearest-neighbor entropy estimator~\citep{Lombardi2016KNNEntropy} as $\delta_i=\log{\left(\frac{1}{N-k}\sum_{j=k}^Nd_{ij}\right)}$, \ie, as the logarithm of its mean distance in parameter space to all other data points $\genome_j$ with a distance $d_{ij}=|\genome_i-\genome_j|$ larger than the $k$-th nearest neighbor distance $d_\mathrm{knn}$ in the dataset buffer.
In practice, we use this diversity metric as a novelty condition $\condition^{(N)}_i=\delta_i$ when conditionally training \CHARLES during the evolutionary process, and sample novel data points with target conditions $\condition^{(NT)}$ that maximize diversity by favoring large $\delta_i$ (see \cref{app:simulation:details:diversity} for details).

We demonstrate this approach on the Rastrigin task discussed in \cref{sub:results:conditional:fitness}, and show the results in \cref{fig:results:conditioning:novelty} (A-D), while a minimal example of the static double-peak environment can be found in \cref{app:results:conditioning:novelty:doublepeak}. We compare fitness and diversity measures of different parametrizations of \HADES and \CHARLES instances, along with mainstream methods such as a simple genetic algorithm~\citep{Ha2017blog} (SimpleGA) and CMA-ES~\citep{hansen2001completely}. Specifically, we utilize the \HADES method as baseline (fitness optimization without diversity condition), and use different combinations of Novelty-based conditioning and Fisher- and greedy  fitness-conditioning (see \cref{sub:results:conditional:fitness}) in different \CHARLES instances, where we use single or multiple conditions during evolutionary optimization (snapshots of converged populations are presented in \cref{fig:results:conditioning:novelty}~(A)).

In our simulations, we start with a narrowly distributed initial condition with \params{$\sigmainit=0.2$}, challenging the respective EAs to explore from a central valley of the fitness landscape and find the different peaks located at $|x|, |y|=3.5$ through exploration (simulation details can be found in \cref{app:simulation:details}). 
We observe from \cref{fig:results:conditioning:novelty} (B) that \HADES and the Novelty-conditional \CHARLES method excel at this task: both methods quickly and reliably identify optimal solutions in \params{$50$} statistically independent optimization runs.
These approaches demonstrate faster convergence than CMA-ES, while the SimpleGA fails to identify the global optima and resides at the nearest locally-optimal peaks to the center.
Moreover, our methods intrinsically maintain a high level of diversity evidenced in the constantly high entropy of the parameters in the population (see \cref{fig:results:conditioning:novelty} (C)).
This diversity manifests through the fact that our methods can identify multiple optima in complex fitness landscapes reliably. We quantify this in \cref{fig:results:conditioning:novelty} (D) where we present the average number of solutions found cumulatively during a single lineage\footnote{We consider a peak in the Rastrigin problem as ``found'' during the course of a single evolutionary search, if at least \params{$10$} independent parameter samples of a single lineage are located within a radius of \params{$0.25$} around this particular peak. Cumulatively found solutions during a particular lineage can thus increase with generations if different peaks are either explored simultaneously or successively by the search procedure.} for all investigated algorithms.
The results show that the vanilla \HADES and the Novelty-conditional \CHARLES can identify \params{$\approx75\%$} of all target peaks reliably, and in \params{$\approx 10\%$} of all simulations, identify all four peaks of the Rastrigin task successfully.

The greedy fitness-conditioning solves the problem more quickly (\cf. inset in \cref{fig:results:conditioning:novelty} (B)) but demonstrates slower convergence on average compared to the previous two solver configurations. The greedy exploration maintains high entropy and, unexpectedly, exhibits explorative behavior over time: as illustrated in \cref{fig:results:conditioning:novelty} (D), the greedy \CHARLES configuration, on average, initially identifies a single peak, yet despite the convergence of average fitness, the number of discovered solutions steadily increases with successive generations.
In contrast, Fisher fitness-conditioning and especially CMA-ES show limited diversity and typically converge onto one optima.

Fisher fitness-conditioning and CMA-ES demonstrate limited diversity and typically converge to a single optimum. Nevertheless, CMA-ES consistently identifies the global optimum in the Rastrigin problem, where the parameter space aligns well with the method's Gaussian generative model~\citep{hansen2001completely}: The algorithm explores the environment by adapting the covariance matrix of a multivariate Gaussian distribution (matching the search space dimension) to best fit the likelihood of the data. The EA's reproduction step is realized by sampling novel data points from this refined generative model, exploring the parameter space through successive Gaussian model adjustments.

In our Rastrigin example, the directions of the nearest local optima near the centered initial generation align with the global optima. Through expansion of the covariance matrix in a particular principal direction, \ie $(x=\pm y)$, CMA-ES efficiently identifies a global optima very efficiently in few generations. 
However, this search strategy exhibits a strong inductive bias and can be inefficient when the search space requires complex reorientations of search directions. 
We propose that deep-learning-based diffusion models provide a more flexible approach to understanding the parameter space due to the universal approximation theorem~\citep{Hornik1989}, reflected in their adaptable search strategy.

To illustrate this, we introduce a ``twisted'' variant of the Rastrigin problem. By transforming the coordinate space $\genome\rightarrow\tilde{\genome}$ non-linearly, we warp the peaks of the Rastrigin function along an outward spiraling pattern, where global maxima are twisted relative to the initial local optima near the center (see \cref{fig:results:conditioning:novelty} (E) and \cref{app:fitness-landscapes:rastrigin} for details).
This geometric modification of the fitness landscape significantly impacts CMA-ES reliability, while our \HADES and \CHARLES methods maintain their performance despite increased complexity (see \cref{fig:results:conditioning:novelty} (E-H)).

In all situations presented in \cref{fig:results:conditioning:novelty}, using the Novelty-condition introduces a repulsive bias between parameter clusters during the DM's generative process.
As illustrated in \cref{fig:results:conditioning:novelty}~(C,G) and (D,H), this leads to two key outcomes: accelerated exploration of high-fitness regions in the parameter space and, simultaneously, to increased population diversity. 
While CMA-ES, SimpleGA, and Fisher-based \CHARLES converge to a single solution with limited diversity, both \HADES and particularly the Novelty-conditional (greedy) \CHARLES solvers maintain significantly higher diversity even after fitness convergence. That shows that a single population is capable of exploring multiple solutions at once, and that the population intrinsically explores new parameter space regions if environmental conditions are changing\footnote{Notably, we utilize a static fitness landscape in \cref{sub:results:conditional:novelty}, but we might consider the population as part of the environment, competing for resources. Novelty conditioning is sensitive to adaptations, especially to clustering of the population in an environment, rendering the environment of \CHARLES as dynamically changing. This, in turn, impacts the reproductive process of the diffusion model across generations, promoting explorative behavior.}.  
On average, the Novelty-conditional \CHARLES method shows increased diversity compared to corresponding non-novelty-conditional configurations.

Conditioning on novelty effectively applies neutral selection pressure, promoting population diversity while independently optimizing fitness scores.
This approach serves as an effective regularization mechanism for DM-evolution, generating diverse and novel solutions while maintaining the ability to exploit clusters of elite solutions.

\subsection{Genetically conditioning behavior: how information traverses scales}
\label{sub:results:conditional:phenotype}
So far, we have demonstrated how the \CHARLES method can be applied to constrain (i) the search dynamics in the parameter-space, (ii) the fitness quality of the samples across generations, and even (iii) improve population wide diversity in a diffusion evolution optimization process.
Building upon these results, we explore its application in selectively sample genotypic parameters to achieve desired phenotypic traits.
Specifically, we explain whether we can conditionally train the DM in \CHARLES using both (i) genotypic representations and (ii) associated phenotypic qualities of agents in Reinforcement Learning (RL) environments~\citep{sutton1998reinforcement}. The goal is to selectively sample RL agents during an evolutionary process that exhibit specific target behaviors, notably without pretraining the DM.

\begin{figure}
    \centering
    \includegraphics[width=\linewidth]{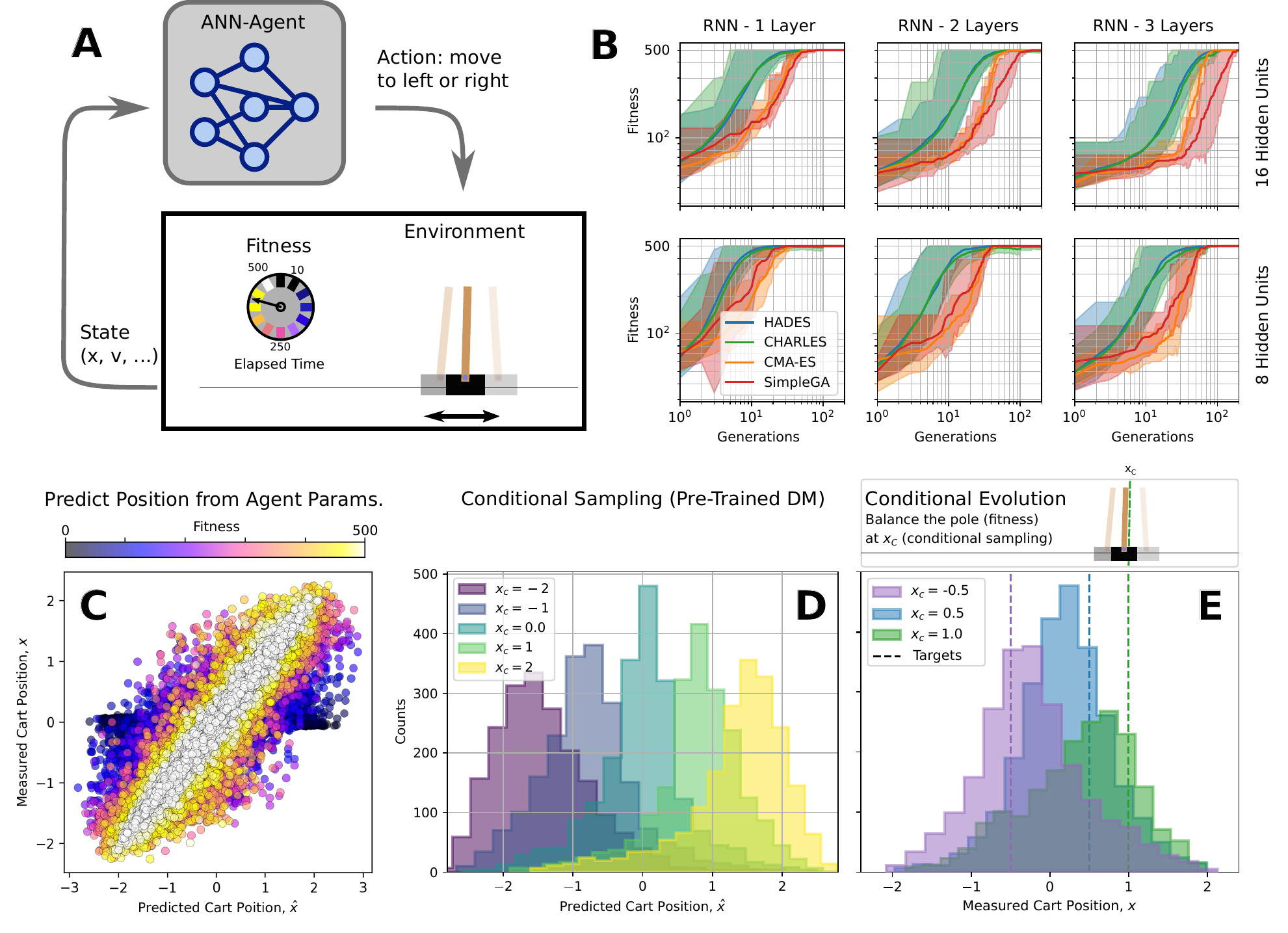}
    \caption{(\textbf{A}) Sketch of an ANN-based RL-agent controlling the cart in a cart-pole environment.
    (\textbf{B}). Training performance of \HADES, Fisher-conditional \CHARLES, CMA-ES, and SimpleGA on the cart-pole task with different ANN architectures. The solid lines emphasize the mean maximum fitness of the different solvers (color-coding) which we averaged over statistically independent evolutionary optimization runs. The shaded areas indicate the best performing individual across independent simulations until a certain generation for larger values than the mean, and the STD of all simulations for lower values than the mean.
    (\textbf{C}) Accuracy of predicting a cart-pole agent's resting position from its ANN parameters (by training a \params{4 layer ANN with of 48 hidden units and LeakyReLU activation} on a dataset of \params{$10$} unbiased evolutionary runs using the \textit{HADES} method); this works well for high-fitness individuals (color-coded).
    (\textbf{D}) Estimated resting position of cart-pole agents sampled conditionally for target resting positions $x_C=\{-2, -1, 0, 1, 2\}$ (color-coding) by a pretrained DM (\params{2 layers with 324 hidden units and ELU activation for 2000 episodes on the same data as in (C)}).
    (\textbf{D}) Resting positions of conditionally evolved lineages with the \CHARLES method that give rise to RL-agent policies with desired behavior of target resting positions \params{$x_C=\{-0.5, 0.5, 1.0\}$} (color coded); the DMs in (E) are not pretrained as in (D), but trained on heuristic data of the respective evolutionary lineage.
    }
    \label{fig:results:conditioning:cartpole}
\end{figure}

Traditional RL applications aim to identify policies that enable autonomous agents to effectively navigate their environments: RL agents perceive different aspects of their environment, such as state information and a reward signal, and need to propose actions that maximize reward acquisition. The policy of an agent, \ie, its internal decision-making machinery, is often modeled by Artificial Neural Networks (ANNs) receiving environmental states as input, and outputting high-quality actions that enable the agent to navigate its environment efficiently. The challenge thus is to identify ANN parameters that enable agents to maximize reward acquisition corresponding to high fitness scores. 
This is often achieved by gradient-based RL algorithms~\citep{schulman2017proximalpolicyoptimizationalgorithms} which require careful curation of differentiable reward signals. Especially in multi-objective scenarios either cumbersome reward-shaping~\citep{Andrew1999PolicyInvarianceRewardTrafo} or curriculum learning techniques~\citep{Bengio2009CurriculumLearning} are necessary to balance different reward signals, or the environment needs to be extremely general~\citep{Silver2021} leading to substantial computational overhead and potentially unpredictable behavior~\citep{Bostrom2013}. In contrast, EAs have proven highly successful to evolve slim and problem-specific ANN-based RL agent policies simply based on cumulative reward measures and often result in much more robust, transferable, and interpretable agent policies~\citep{hartl2024neuroevolutiondecentralizeddecisionmakingnbead, hartl2024evolutionary, Hartl2021, Tang2021, Tang2020, Ha2017blog, Stanley2002}. As demonstrated in \cref{fig:results:conditioning:cartpole}, \HADES proves highly effective for this purpose.

The goal of our research extends beyond evolving RL agents with high fitness. We seek to develop agents that exhibit specific target behaviors not encoded in the environment's reward signal, thus remaining neutral to the agent's fitness. Specifically, we apply classifier-free-guidance at the genotypic level of \CHARLES to evolve RL-agents with targeted phenotypic-behavioral traits in their respective environment.

To validate our approach on a minimal yet descriptive example, we employ the cart-pole system~\citep{barto1983neuronlike}. In this system, a cart with a hinged pole moves sideways with the objective of maintaining the pole in a vertical position for as long as possible within a defined range, as illustrated in \cref{fig:results:conditioning:cartpole}~(A).
The fitness score corresponds to the total number of time steps $N_s\leq500$ before the game termination, which occurs when either the pole-angle exceeding $\phi_\mathrm{lim}=\pm 12^{\circ}$ or the cart moving beyond the boundaries $x_\mathrm{lim}=\pm 2.4$.
The cart is controlled by an ANN (detailed in \cref{app:agents:cartpole}). The network processes four input parameters, \ie the current position, $x(s)$, velocity $\dot x(s)$, pole angle $\phi(s)$, and pole angular velocity $\dot\phi(s)$, and its outputs determines the cart's movement direction (left or right). The task is considered solved if the RL agent consistently achieves a fitness score of $500$ across multiple episodes.

First, we evolve RL-agents with maximum fitness using \HADES, comparing its performance against mainstream methods such as CMA-ES and a SimpleGA. As ANN architecture, we chose Recurrent Neural Networks (RNN) with a different number of layers and hidden units. The results depicted in \cref{fig:results:conditioning:cartpole}~(B), demonstrate, that the \HADES method is capable of solving the problem in as little as \params{$3-4$} generations, although RNNs can be tedious to train in RL applications, while both CMA-ES and the SimpleGA take longer by an order of magnitude; we used a population size of \params{$\popsize=256$} and fitness scores averaged over \params{$N_e=16$} episodes.  Notably, every implementation utilizes elitism. The absence of this feature would further widen the performance gap between our methods CMA-ES and SimpleGA method.
We expanded our comparison to include an analysis of the \CHARLES method, method, specifically when conditioning on producing offspring with high fitness. The results demonstrate that even when operating under Fisher conditional optimization, the performance noticeably exceeds that of the \HADES. Notably, every implementation features elitism. The absence of this feature would further widen the performance gap between our methods CMA-ES and SimpleGA would be even more significant.

Next, we focus on controlling the \textit{resting position} $x^{(r)}=\frac{1}{N_s}\sum_{s=N_s-N_r}^{N_s}x(s)$, which expresses the average position where the cart stabilizes the pole during the final \params{$N_r=100$} steps of an episode. We conduct \params{$10$} independent \HADES optimization trials, during which we recorded both the ANN parameters and their corresponding behavioral outcomes $\{\genome_i, x_i^{(r)}\}$.
To predict the resting position based on the RL-agent parameters, we train a \params{deep ANN} denoted as $f^{(r)}$, such that $\hat{x}^{(r)}_i=f^{(r)}(\genome_i)$. We aim to minimize the weighted mean-square error $h[f_i]|\hat{x}^{(r)}_i - x^{(r)}_i|^2$, where $h[f_i]$ serves as the weighting factor (\cf, \cref{sub:methods:DM} and \cref{app:fitness:weighting:roulette}). As shown in \cref{fig:results:conditioning:cartpole}~(C), the prediction accuracy is particularly strong for agents with high fitness of $\approx f_i\approx500$. Notably, these high-fitness agents demonstrate resting positions that span the complete range of $x\in[-2,2]$. 
To predict the resting position from the RL-agent parameters, we train a deep ANN, $f^{(r)}$, such that $\hat{x}^{(r)}_i=f^{(r)}(\genome_i)$. The training minimizes the mean-square error $h[f_i]|\hat{x}^{(r)}_i - x^{(r)}_i|^2$, where $h[f_i]$ serves as the weighting factor (detailed in \cref{app:fitness:weighting:roulette}). As shown in \cref{fig:results:conditioning:cartpole}~(C), the prediction accuracy is particularly strong for high-fitness agents with $f_i\approx500$.
These high-fitness agents demonstrate resting positions spanning the complete range $x\in[-2,2]$.

Using the joint database of parameters and associated behavioral data $\{\genome_i, x_i^{(r)}\}$ from earlier \HADES evaluations, we \params{train a generative DM} ``offline'' on this genetic database, \ie, without utilizing the DM in any further optimization.
We then use the DM to conditionally sample novel genotypes $\hat{\genome}_\nu$ that parameterize the behavior of RL-agents to balance the pole at specific target locations \params{$x_c=\{-2, -1, 0, 1, 2\}$}. The results in \cref{fig:results:conditioning:cartpole}~(D) demonstrate, that the conditional sampling achieves good accuracy in biasing the corresponding RL-agents towards exhibiting the desired target behavior. 

Finally, we use the \CHARLES method to evolve an initially randomized population of ANN parameters for cart-pole RL-agent controllers toward solutions that exhibit specific target behaviors, namely balancing the pole either at resting positions \params{$x_c=\{-0.5, 0.5, 1\}$}.
This is achieved without modifying the reward signal, but instead using conditional training of the generative DM on the ANN parameters $\genome_i$ and their associated mean resting positions $x_i^{(r)}$ measured during the evolutionary process. Offspring genotypes are then selectively generated by conditioning the DM on $\condition^{(r)}=x_c$ (see \cref{app:agents:cartpole:conditioning} for details).

Notably, this experiment is performed independently of the database and pretrained DM discussed in \cref{fig:results:conditioning:cartpole} (C, D). Rather, we begin with a randomized initial population and a randomly initialized DM, that is solely on parameters explored during the respectively evolving populations of cart-pole agents, \ie, in an ``online'' mode.
\Cref{fig:results:conditioning:cartpole}~(E) shows the measured resting positions $x_i^{(r)}$  for the \params{three} independent lineages with \params{$x_c=-0.5, 0.5$ or $ 1$}. These positions $x_i^{(r)}$ correspond to generated agents $\genome_i$ that achieved an average fitness score of $500$ over \params{$N_e=16$} consecutive episodes.

The generative process demonstrates clear bias in generating genotypic parameters that encode specific functional behaviors at the phenotypic level that is neutral to the agents' fitness scores.
Beyond intriguing potential biological implications (discussed below), our \CHARLES method enables efficient multi-objective optimization across various problem domains without requiring complex reward shaping techniques~\citep{Andrew1999PolicyInvarianceRewardTrafo}.
Through classifier-free-guidance conditioning of the generative process at genotypic parameter level, we direct evolutionary search toward parameter space regions that yield desired phenotypical behaviors.
\CHARLES effectively channels conditional information (specifically, desired behavior) across multiple layers of abstraction, from the generative phase at the genotypic level to actual behavior of ANN-based autonomous RL-agents at the phenotypic level in physically realistic environments. This bridging occurs through joint training of a generative model on governing genotypic parameters and their associated phenotypic traits. DMs capture the developmental layer between genotypes and phenotypes~\citep{levin2023darwin} by learning how to gradually generate parameters, notably through step-wise denoising operations, that conform to conditionally encoded phenotypic traits. In essence, DMs encapsulate the computationally irreducible nature~\citep{Wolfram2002} of the developmental process by learning how to actually compute high-quality samples through step-wise error correction mechanisms conditional to specified target features, thereby forming an associative memory of domain-specific generative processes.

\section{Discussion}
\label{sec:discussion}
Our work, alongside a complementary contribution~\citep{zhang2024diffevo}, establishes a connection between diffusion models (DMs) and evolutionary algorithms (EAs) through shared underlying conceptual and mathematical foundations: DMs can be viewed as evolutionary processes in disguise. 
In this paper, we demonstrate that deep-learning based DMs can effectively serve as efficient generative models in EAs, enhancing genotypic recombination operations. 
Rather than relying on pretraining DMs with large general datasets, we continuously refine DMs using heuristically acquired, high-quality parameters from evolutionary process. This iterative refinement of the DM's generative process, based on the most recent evolutionary evidence in biology, enables the DM to be adaptive to evolutionary changes.

DMs leverage Artificial Neural Network (ANN) to sample novel data points conforming to a target parameter distributions. Given their status as universal function approximators~\citep{Hornik1989}, ANNs excel at learning complex correlations within arbitrary datasets, making them ideal for identifying subtle correlations in parameters of evolutionary processes.
Through iterative refinement of the DM's generative process using heuristically acquired high-quality data from evolutionary processes, we propose the novel \textit{Heuristically Adaptive Diffusion-Model Evolutionary Strategy} (\HADES) method.
We contrast our method's performance with mainstream EA techniques across various numerical optimization scenarios and report significant improvements in adaptability to changing environments while maintaining reliable convergence to target solutions.

DMs augment evolutionary processes with unconventional (epigenetic) memory~\citep{jablonka_2009_transgenerational, jablonka_2017_the}: Using elite buffer datasets collected across generations or persistent traits in constantly retrained DMs allows the generative process to utilize previously experienced information, thus enabling faster adaptation in changing environments.
This memory capability proves especially crucial when objective functions are computationally expensive. Our findings confirm that maintaining a memory buffer enhances both search result quality and diversity.

Moreover, via classifier-free-guidance techniques~\citep{ho2022classifier}, we can utilize conditional sampling in DMs to directly bias the evolutionary search dynamics, steering it towards regions in the parameter space that exhibit desired target traits. This leverages multi-objective optimization without the need for complex reward shaping~\citep{Andrew1999PolicyInvarianceRewardTrafo} or curricula learning techniques~\citep{Bengio2009CurriculumLearning}.
By conditioning the DM's generative phase across successive generations, we demonstrate control over: 
(i) the search dynamics in the parameter-space, 
(ii) sampling offspring generations with specific fitness distributions similar to inverse reinforcement learning~\citep{schmidhuber2020reinforcementlearningupsidedown} and in addition controlling the diversity or greediness of the population, 
(iii) biasing the search towards desired phenotypic traits that are neutral to the problem's fitness score, and 
(iv) even to explicitly maintain population diversity through novelty- or diversity-conditioning.
This conditional sampling mirrors image- or video-generation techniques that are controlled by custom text-inputs.
Our \textit{Conditional, Heuristically-Adaptive ReguLarized Evolutionary Strategy through Diffusion} (\CHARLES) method introduces conditional sampling during optimization, effectively constraining search result qualities via custom control parameters defined independently to the task's objective function. 
Thus, our approach represents, to the best of our knowledge, the first  ``Talk to your Optimizer''~\citep{Mathews2018} application.

In that way, DMs demonstratively outperform the generative models of mainstream genetic algorithms in flexibility, versatility, and control over the search dynamics:
With \HADES and \CHARLES we can successfully solve high-dimensional complex optimization problems, and even bias the search-dynamics toward desired behavior, out-competing mainstream approaches.
Through an epigenetic dataset memory buffer, we can even dynamically condition the search behavior to revisit promising, previously experienced parameter space regions, similar to an associative memory~\citep{Ambrogioni2023, Watson2014, power2015can}.
Moreover, such unconventional memory properties enable specific conditioning of the generative process across evolutionary generations to actively promote novelty and diversity, as effective novelty search requires memory capability~\citep{stanley2015greatness}.
Eventually, both \HADES and \CHARLES solutions show remarkable diversity, particularly in reinforcement learning tasks.

Several limitations of the current approach can be extended  in future work:
First, while biological evolution operates with discrete chemical building blocks, we have applied our algorithm only to continuous parameter spaces.
Although we utilized several DM architectures and observed largely architecture agnostic search dynamics with sufficiently large DM latent spaces, the balance between population size, memory buffer dataset size, training epochs, learning rate schedule, and the number of evolutionary generations between DM retraining may affect search dynamics.
High quality data proves crucial for both diversity and convergence rates.
Careful consideration is necessary to balance exploration-exploitation behavior, particularly regarding result diversity. Questions remain about whether to retrain a single DM between generations or train different DMs from scratch using updated heuristic data.
Moreover, the ratio of the active population to the total possible solutions in the parameter space needs to be explored.

Intriguingly, our methods demonstrate several aspects relevant to the emerging field of Diverse Intelligence ~\citep{levin_technological_2022}, in which cognitive and epistemic dynamics are studied in a very wide range of embodiments and spatio-temporal scales. For example, the DM's generative process applies successive denoising steps, transforming initially random input into successively refined highly-correlated, high-quality parameter output conforming to a training dataset.
This iterative denoising process represents a perception-action cycle of genotypic parameters similar to that studied in the behavior of agents~\citep{Vernon_EmbodiedCognition, GORDON_The_road_towards_understanding} and in Neural Cellular Automata~\citep{Mordvintsev2020, Li2002}, driving (\ie, poorly adapted) genotypes towards statistically more probable (\ie, better adapted) parameter space regions.
Furthermore, this process can demonstratively learn from experience and, through conditioning, responds to external stimuli that are orthogonal, or neutral to a fitness score.
This perspective reframes the evolutionary process as an active learning system~\citep{watson2016can}. 
Thus, an evolutionary process's classification as being a variational or transformational depends on the observer's perspective: individuals experience learning as transformational, refining an internal world model while maintaining their ``identity''. From an evolutionary or ecosystem perspective~\citep{VariationalEcology_and_the_physics}, individuals represent temporary ``experiments'', while interacting species from a transformational learning system conditional to particular environmental constraints~\citep{watson2023collective,watson2022design,kouvaris2017evolution,watson2016can,watson2016evolutionary,power2015can,Watson2014}. 

\subsection*{A paradigm shift: Diffusion Models sample Gene-Expressions rather than Genomic Parameters}
Diffusion models have recently been identified as associative memories~\citep{Ambrogioni2023, Hopfield1982}, and, in a complementary work, as evolutionary algorithms~\citep{zhang2024diffevo}.
Correspondingly, evolution can be understood as a form of Hebbian-learning~\citep{watson2023collective, kouvaris2017evolution, watson2016can, watson2016evolutionary, power2015can, Watson2014}.
Evolutionary learning operating at the bio-molecular level, maintaining DNA-based associative memory with self-regulatory capabilities in expressing protein sequences that constrain functionality of their host cell. This perspective presents DNA as a generative model that initializes the developmental process of an organism rather than a direct blue-print of the latter~\citep{mitchell2024genomic, levin2023darwin, Friston2023Variational}. In this process, genes are expressed in modular response to host-cells configuration, internal state, and environment.

Our research demonstrated that DMs, when trained on generating functional parameters conditionally to associated behavioral features, can be used to selectively evolve agent policies exhibiting targeted behavior that is neutral to the fitness score. We explicitly show this in experiments with a cart-pole environment where agents balance a pole vertically, but conditional at specific target locations: The same DM model can generate distinct control parameters for a cart-pole agent, enabling pole stabilization at different desired locations, \eg, $x_A$, or $x_B\neq x_A$, simply by conditioning the DM's generative denoising process on policy $A$ or $B$.
Moreover, we can smoothly transform between these policies by exchanging the respective conditions: When shifting from policy $A$ to $B$, the DM adapts the agent's controller's parameters accordingly, representing an unconventional form of behavioral control through parameter reconfiguration. This mechanism closely parallels how gene-regulatory networks dynamically reconfigure cell functionality in response to internal or external stimuli~\citep{jaeger_bioattractors_2014, alon_introduction_2006, alon_network_2007, davidson_gene_2006, levine_gene_2005, kauffman_metabolic_1969}.

Viewing this through the lens of recent interpretations of the genome as a generative model~\citep{mitchell2024genomic, levin2023darwin, Friston2023Variational}, we hypothesize that the DM used in our methods literally represents a lineage's evolving genome, including its ability to self-regularize and utilize gene expressions that reconfigure phenotype functionality based on environmental constraints (as demonstrated by conditions $A$ or $B$ in the above cart-pole example). This suggests that the parameters in our work do not represent  genotypic representations but gene expressions encoding phenotype functionality and behavior, similar to the ways in which gene expressions encode protein sequences that control cellular behavior. The complete genome (\ie, represent by the DM) contains much richer information, including an associative memory that is accessible when needed.
We propose that generative diffusion models are good models for self-regulatory, generative DNA, sampling gene-expressions rather than genotypic representations

Thus, we believe that our model is much closer to biology than previous evolutionary methods, by representing DNA as a generative model with associative memory, surpassing previous evolutionary methods in detail and functionality. This representation enables our evolutionary search strategy to specifically respond to external conditions and generates problem-specific parameter expressions through conditional denoising processes that are fundamentally rooted in non-equilibrium physics~\citep{sohl2015deep}. Such capabilities allow dynamic reconfiguration of phenotype behavior, closely mirroring biological systems' adaptability, in turn promoting intriguing technological innovations.

\section*{Acknowledgements}
We thank Sebastian Risi, Léo Pio-Lopez, Patrick Erickson, Josef Kaufmann, and Lukas Rammelmüller for helpful discussions. 
BH acknowledges an APART-MINT Stipend of the Austrian Academy of Sciences. 
The authors acknowledge the Tufts University High Performance Compute Cluster~\footnote{\url{https://it.tufts.edu/high-performance-computing}} and the Vienna Scientific Cluster~\footnote{\url{https://www.vsc.ac.at}} which have been utilized for the research reported in this paper.
ML gratefully acknowledges support via Grant 62212 from the John Templeton Foundation.

\section*{Author declarations section}
The authors have no conflicts to disclose.

\section*{Data availability statement}
Computational protocols and numerical data that support the findings of this study are shown in this article, and the Appendix. 


\bibliography{manuscript.bib}
\bibliographystyle{plainnat}  

\newpage
\appendix
\FloatBarrier

\section{Algorithmic Details}
\subsection{Fitness re-weighting with the Roulette-Wheel Method}
\label{app:fitness:weighting:roulette}
In order to reweigh the importance of parameters $\genome_i$ from the training dataset $\dataset=\{\genome_1,\genome_2,\dotsc,\genome_i,\dotsc,\genome_{\popsize}\}$ during training of the diffusion models, we rely on a reweighing function $h[f_i]$ of the parameter fitness $f_i=f(\genome_i)$. 
Inspired by ``roulette-wheel selection''~\citep{holland1992adaptation}, we define the remapping function $h[f_i]$ assuming sorted fitness scores, $f_i\leq f_{i+1}$, as
\begin{equation}
    h[f_i; w] = \frac{w\,\sum_{j=1}^i F(f_j)}{\sum_{k=1}^{\popsize}\sum_{j=1}^k F(f_j)},
    \label{eq:methods:roulette}
\end{equation}
where we introduced the relative fitness 
$F(f_i)=\exp{\left(s(f_i-f_{\min})/(f_i-f_{\max})\right)}$, with selection pressure $s$. $f_{\min}$ and $f_{\max}$ are the minimum and maximum fitness values associated to the elements in current population $\population_\gt=\{\genome_{\gt,i}\}$.
The weighting factor $w$ is either set to $w_\mathcal{N}=1$, leading to a normalized $h[f_i]$, or represents the fitness weights before rescaling $w_f=\sum_{i=1}^{\popsize}|f_i|$. 

In practice, we follow two paths: (i) we either reweigh the parameter loss defined in \cref{eq:methods:dm:loss} via a reweighing function $h[f_i, w_f]$, or (ii) utilize the reweighing function $h[f_i, w_\mathcal{N}]$ as a probability for selecting training data points from the dataset buffer $\dataset$.

\subsection{Blending Generative- and Heuristic Crossover with Mutations}
\label{app:hades:sampling}
To generate a novel generation $\population_{\gt+1}=\{\genome_{\gt+1,i}\}$, we either rely on the \textbf{generative process} $\genome_{\gt+1,j}\sim p(x)$ of the diffusion model $\mathcal{G}$, or perform \textbf{manual crossover} operations $\genome_{\gt+1,k}=\genome_{\gt,l}\bigoplus\genome_{\gt,m}$ of existing genetic material in the current population $\population_\gt$; for the manual crossover, we select two parent genotypes $\genome_{\gt,l},\genome_{\gt,m}\sim h[f_e; w_\mathcal{N}]$ from the $N_e$ elite solutions\footnote{Notably, we do not use elitism in our approach, but use elite solutions to sharpen different selection criteria.} with fitness $f_e$ larger than $(\popsize-N_e)/\popsize$ percent of the current population $\population_{\gt}$ (see \cref{eq:methods:roulette}), and chose parameters of either parent $l, m$ at random to form an offspring genotype $\genome_{\gt+1,k}$.
We sample a total number of $(\popsize - N_c)$ novel parameters from the diffusion model, and perform $N_c$ manual crossover operations; thus, the parameter $N_c\in[0,\popsize]$ is termed \textit{crossover ratio}.

Additionally, we employ \textbf{mutation operations} manually for both parameters sampled by the diffusion model and from manual crossover.
To this end, we define a \textit{mutation scale} parameter $t_\mu$, and fully rely on the diffusion operation defined in \cref{eq:methods:dm:diffuse} to mutate a parameter $\genome\rightarrow\sqrt{\alpha_{t_\mu}}\genome + \sqrt{1-\alpha_{t_\mu}}\bm\epsilon$, with $\bm\epsilon\sim\mathcal{N}(0,I^D)$.
Moreover, we define a \textit{mutation ratio} $N_\mu\in[0,\popsize]$ specifying the fraction of the novel generation $\population_{\gt+1}$ that is randomly selected to be subjected to mutation via $t_\mu$ diffusion steps.

We allow ``readaption'' of such noisy parameters at a \textit{readaptation rate} $t_a$: to this end, we use the novel and mutated population as initial configurations for the diffusion model and perform $t_a$ denoising steps, \ie, starting at diffusion-time $t=t_a$; Notably, $\vx_{t=0}$ represent fully denoised samples while $\vx_{t=T}$ samples are fully diffused.

\subsection{Symbols}
All symbols used in our methods are collected in \cref{tab:symbols}.

{\renewcommand{\arraystretch}{1.5}
\begin{xltabular}{\linewidth}{lp{5cm}X}
\caption{List of Symbols and Notations\label{tab:symbols}} \\
\hline
\textbf{Symbol} & \textbf{Definition} & \textbf{Example/Usage} \\
\hline
\endfirsthead

\multicolumn{3}{l}{\tablename\ \thetable{} -- continued from previous page} \\
\hline
\textbf{Symbol} & \textbf{Definition} & \textbf{Example/Usage} \\
\hline
\endhead

\hline \multicolumn{3}{r}{\textit{Continued on next page}} \\
\endfoot

\hline
\endlastfoot
\multicolumn{3}{l}{\textit{General Parameters}} \\
$D$ & Parameter dimension & Dimension of search space \\
$\gt$& Generation index & $\gt = 1,2,\ldots,N_\gt$ \\
$N_\gt$ & Total number of generations & Evolution stops at $\gt = N_\gt$ \\
$N_p$ & Population size & Number of individuals per generation \\
$N_e$ & Elite ratio & Number of best performing individuals from population \\
$N_c$ & Crossover ratio & Number of individuals from population for crossover \\
$N_\mu$ & Mutation ratio & Number of individuals from population for mutation \\
$t_\mu$ & Mutation scale & Number of diffusion time-steps used to mutate samples \\
$t_a$ & Readaptation rate & Number of diffusion time-steps to readapt (de mutated samples \\

$\sigma_I$ & Initial standard deviation & STD of initial population \\
$s$ & Selection pressure & Sharpness of ``roulette wheel selection`` fitness reweighing in \cref{eq:methods:roulette}\\
\hline
\multicolumn{3}{l}{\textit{Population and Genome}} \\
$\genome$& Genome variables & Individual parameter set \\
$\population$& Population & Set of all genomes in a generation \\
$\population_\gt$& Population at generation $\gt$ & $\{\population_{\gt,1}, \genome_{\gt,2}, \ldots, \genome_{\gt,N_p}\}$ \\
$\dataset$ & Dataset buffer & Collection of past elite solutions \\
\hline
\multicolumn{3}{l}{\textit{Diffusion Model Parameters}} \\
$t$ & Diffusion time & $t = 1,2,\ldots,T$ \\
$\vx_t$ & State at diffusion time $t$ & Parameters during diffusion process \\
$\alpha_t$ & Noise schedule & Controls noise level at time $t$ \\
$\ve$ & Random noise & $\ve \sim \mathcal{N}(0,I^D)$ \\
$\epsilon_\theta$ & Neural network & Predicts noise during denoising \\
$\mathcal{G}$ & Generative model & Maps current to next generation \\
\hline
\multicolumn{3}{l}{\textit{Fitness and Conditioning}} \\
$f(\genome)$ & Fitness function & Evaluates quality of genome \\
$h[f]$& Weighting function & Maps fitness to sampling probability \\
$\condition$& Condition variables & Target for conditioning \\
$\ceval(\genome)$& Classifier function & Maps genome to condition \\
\hline
\multicolumn{3}{l}{\textit{Diffusion Model Architecture (typically feed-forward)}} \\
$N_\mathcal{L}$ & Number of hidden layers & - \\
$N_\mathcal{H}$ & Number of hidden units per hidden layers & - \\
$f_\mathcal{F}$ & Activation function of hidden units & - \\
$\lambda_\mathrm{LR}$ & Learning Rate & The learning rate used during training. \\
$N_\mathcal{E}$ & Epochs & The number of training epochs. \\
\hline
\multicolumn{3}{l}{\textit{Special Functions}} \\
$p(\vx)$ & Probability density & Distribution of parameters \\
$L(\theta)$ & Loss function & Training objective for diffusion model \\
\hline
\end{xltabular}
}

\FloatBarrier

\section{Simulation Details}
\label{app:simulation:details}
\FloatBarrier
\Cref{app:tab:simulation:details} provides an overview of the simulation parameters for the results depicted in the main text. Additional remarks can be found in the text below.

\begin{xltabular}{\linewidth}{l|c|c|c|c|c|c|c|c}
\caption{\HADES and \CHARLES simulation parameters for the results presented in the main text. See \cref{tab:symbols} for more details about the symbols.}
\label{app:tab:simulation:details} \\
\hline
\textbf{Experiment} & \textbf{Solver} & $\popsize$ & $\sigmainit$ & $N_\mathcal{B}/\popsize$ & $N_e/\popsize$ & $N_c/\popsize$ & $N_\mu/\popsize$ & $t_\mu / T$ \\
\hline
\cref{fig:results:dynamic:env} & \HADES & 256 & 0.5 & 1 & $0.15$ & 0 & 1 & $5\times10^{-2}$ \\
\cref{fig:results:conditional:genotype} & \CHARLES & 256 & 2 & 3 & $0.15$ & $2^{-3}$ & 1 & $5\times10^{-2}$ \\
\cref{fig:results:conditioning:dynamic} & \CHARLES & 256 & 2 & 3 & $0.15$ & $2^{-3}$ & 1 & $5\times10^{-2}$ \\
\cref{fig:results:conditioning:novelty} & \CHARLES & 256 & 0.2 & 10 & $0.25$ & $0$ & 0.1 & $2\times10^{-1}$ \\
\cref{fig:results:conditioning:cartpole} (B) & \HADES & 256 & 0.5 & 4 & 0.2 & 0.4 & 1 & 0.1 \\
\cref{fig:results:conditioning:cartpole} (C) & \HADES & 256 & 0.5 & 4 & 0.2 & 0.4 & 1 & 0.1 \\
\cref{fig:results:conditioning:cartpole} (D) & \CHARLES & 32 & 0.5 & 16 & 0.25 & 0.45 & 0.1 & 0.2 \\
[1ex]
\hline
\multicolumn{9}{c}{} \\[-2ex]
\hline
\textbf{Experiment} & $t_a$ & $s$ & $w$ & $N_\mathcal{L}$ & $N_\mathcal{H}$ & $f_\mathcal{F}$ & $\lambda_\mathrm{LR}$ & $N_\mathcal{E}$ \\
\hline
\cref{fig:results:dynamic:env} & 0 & 10 & $w_\mathcal{N}$ & 3 & 24 & Leaky-ReLU & $10^{-3}$ & 100 \\
\cref{fig:results:conditional:genotype} & 0 & 5 & $w_\mathcal{N}$ & 3 & 24 & Leaky-ReLU & $10^{-2}$ & 200 \\
\cref{fig:results:conditioning:dynamic} & 0 & 5 & $w_\mathcal{N}$ & 3 & 24 & Leaky-ReLU & $10^{-2}$ & 200 \\
\cref{fig:results:conditioning:novelty} & 0 & 12 & $w_f$ & 2 & 64 & ReLU & $10^{-2}$ & 200 \\
\cref{fig:results:conditioning:cartpole} (B) & 0 & 18 & $w_\mathcal{N}$ & 4 & 32 & Leaky-ReLU & $10^{-2}$ & 200 \\
\cref{fig:results:conditioning:cartpole} (C) & 0 & 18 & $w_\mathcal{N}$ & 4 & 32 & Leaky-ReLU & $10^{-2}$ & 200 \\
\cref{fig:results:conditioning:cartpole} (D) & 0 & 8 & $w_f$ & 3 & 324 & ELU & $10^{-3}$ & 500 \\
\hline
\end{xltabular}

For training the diffusion model for \HADES and \CHARLES experiments, we use the \textit{Adam} optimizer, a batch-size of 256, and an $L2$ weight-decay of $\lambda_{L2}=10^{-5}$ in all our simulations.

For the results depicted in \cref{fig:results:conditioning:novelty}, we maintain a population-size of \params{$\popsize=256$}, \params{$k=128$, $\beta=10$, and $\Delta=10^{-8}$}.

\section{Sampling Diversity Conditions}
\label{app:simulation:details:diversity}
During sampling of the diffusion model in \cref{sub:results:conditional:novelty}, we need to identify target conditions $\condition^{(NT)}$ that bias the generative process to bring forth novel data points $\genome_\nu$ with large diversity measure. These conditions are related to entropy measures~\citep{Lombardi2016KNNEntropy} of the current population and don't have a well-defined objective target value (the diversity measure is a dynamic property of an individual in a given population rather than an objective trait at the individual level). 
Thus, we sample the target conditions $\condition^{(NT)}\sim p_N$ at every generation from a heuristic Boltzmann distribution $p_N\propto\exp{\left(-\beta E(f, \condition^{(N)})\right)}$, with $E(f_i, \condition^{(N)}) = \frac{\tilde{f_i}}{\condition^{(N)}_i+\delta_0}$, where we introduced the rescaled fitness $\tilde{f}_i=1-\frac{f_i - f_{\min{}}}{f_{\max{}}-f_{\min{}}}$, and transform $\condition^{(N)}_i$ to positive values by adding $\delta_0=|\min_i(\delta_i)+\Delta|$.
In that way, target conditions with large $\delta_i$ are sampled with higher probability, which effectively biases the generative process of the diffusion model towards sampling novel offspring genotypes with larger diversity $\delta_\nu\leq\max_i{(\delta_i)}$ than present in the current population, $\delta_i$.

\section{Two-Dimensional Fitness Landscapes}
\label{app:fitness-landscapes}
\subsection{Fitness- and Diversity Conditioning for the Double-Peak Task}
\label{app:results:conditioning:novelty:doublepeak}
We used the double-peak problem as minimal toy-example for the results shown in the main text. We refer to \cref{eq:doublepeak:dynamic} for the definition and the main text for details.

\begin{figure}
    \centering
    \includegraphics[width=0.7\linewidth]{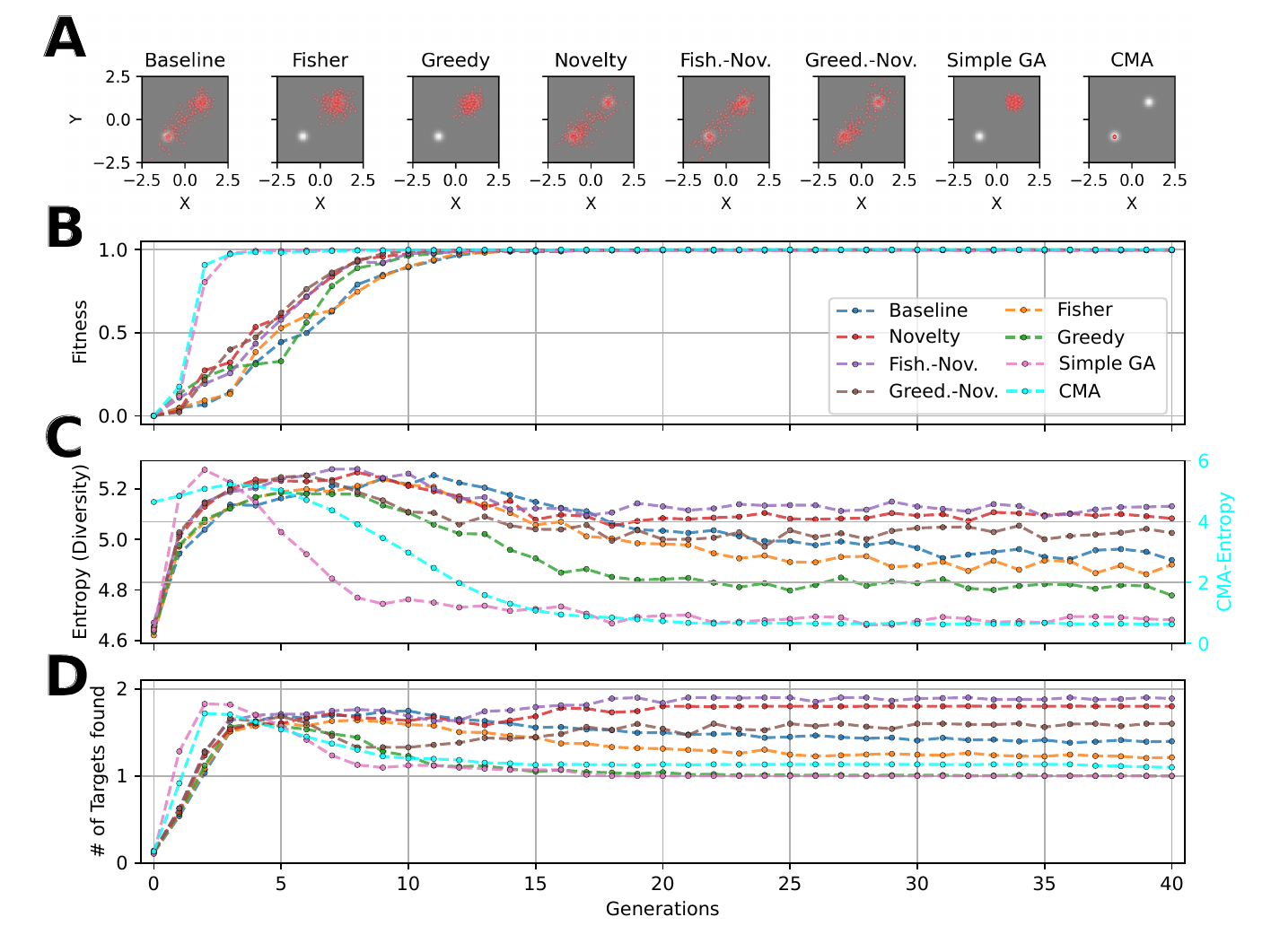}
    \caption{Fitness- and Diversity Benchmarks for the Double-Peak Task.
    (\textbf{A}) Exemplary populations after \params{$40$} generations in the 2D double-peak fitness landscape, see \cref{eq:doublepeak:dynamic}, for different configurations of the \HADES and \CHARLES method contrasted with SimpleGA and CMA-ES algorithms (see text).
    (\textbf{B}) Maximum fitness for the different solver configurations illustrated in (A), 
    (\textbf{C}) entropy-based diversity (see text) of the population, and 
    (\textbf{D}) number of identified targets in the 2D double-peak fitness landscape vs. generations, 
    averaged  over \params{15} statistically independent simulations, respectively.
    The novelty-conditional \CHARLES method reliably maintains diverse populations of high-quality genetic material, demonstrating that our approach reliably identifies multiple optima through conditional diversification.
    }
    \label{fig:results:conditioning:novelty:doublepeak}
\end{figure}

Similar to \cref{fig:results:conditioning:novelty}, we here discuss the results of different \HADES and \CHARLES configurations on the minimal double-peak task. The results are shown in \cref{fig:results:conditioning:novelty:doublepeak}. While for this particular situation, CMA-ES and the SimpleGA converge the fastest (see averaged fitness dynamics in \cref{fig:results:conditioning:novelty}~(B)), the Novelty-conditional \CHARLES instances are second in performance, even better than the fitness-conditional and the baseline algorithms. This is rooted in the fact that the fitness signal of the narrow initial population is small, and all \HADES and \CHARLES instances are used in a non-greedy setting with low selection pressure (see \cref{app:simulation:details}).

Moreover, we see that the vast majority of the Novelty-conditional \CHARLES instances (but also to some extend the baseline \HADES solver) reliable identify and even stabilize on both peaks in the fitness landscape, see \cref{fig:results:conditioning:novelty}~(D), while the more greedy fitness-conditional algorithms often collapse onto one solution after many generations.
In contrast, the state-of-the-art CMA-ES and SimpleGA methods basically zoom in onto one peak very quickly.

Notably, in contrast to the results discussed in \cref{sub:results:conditional:novelty}, we don't use any form of mutational exploration in this setting.

\subsection{The inverse and twisted Rastrigin Task}
\label{app:fitness-landscapes:rastrigin}
The Rastrigin function is defined by $f(\textbf{x})=An+ \sum_{i=1}^2 [ x_i^2 -A \cos (2 \pi x_i )]$ with $\textbf{x}=(x_1, x_2)$; we chose $A=10$.
For the ``twisted''-Rastrigin function, we use a spiral coordinate transformation $\textbf{x}\rightarrow \tilde{\textbf{x}}=r\times(\cos(\theta+\omega r), \sin(\theta+\omega r))$, with polar coordinates $r=|\mathbf{x}|$ and $\theta=\mathrm{atan2}(x_2, x1)$, and constant $\omega$. We evaluate $h(\textbf{x})=f(\tilde{\textbf{x}})$.
For both the Rastrigin and twisted-Rastrigin function, we evaluate use the negative function value for our optimization experiments truncate the function values for $|x_i|>4$ to $0$. In that way, we establish an oscillatory fitness landscape with four maxima located at $|x_i^*|=3.5$ for the Rastrigin function, with a maximum fitness value of $f_\mathrm{max}\approx 64.625$. The maxima of the twisted-Rastrigin function are correspondingly transformed.

\subsection{Grid-Based Entropy Estimate}
\label{app:entropy:estimate}

Similar to Ref.~\citep{zhang2024diffevo}, we quantify the diversity of the solutions by 
(i) dividing the 2-D plane into a grid of $101\times 101$ in the range $x,y\in[-6,6]$, and
(ii) counting the frequencies of all solutions of a particular generation falling into different grid cells $i$. The entropy $H$ is then evaluated as
\begin{equation}
    H=\sum_{i=1}^N P_i\log_2 P_i,
\end{equation}
where $P_i$ is the probability of data points being located in grid $i$. This simple and coarse method allows us to quantify entropy focusing solely on the diversity of solutions across different basins and explicitly avoiding contribution of local diversities. 

\section{Cart-Pole Agents}
\label{app:agents:cartpole}
In \cref{fig:results:conditioning:cartpole}, we utilized different Artificial Neural Network (ANN) architectures to benchmark our algorithms.
More specifically, we used multilayer feed-forward (FF), recurrent-~\citep{Rumelhart1986a} (RNN), and recurrent gene-regulatory networks~\citep{hartl2024evolutionary} (RGRN).

For the results depicted in \cref{fig:results:conditioning:cartpole}~(B), we used RNNs with one to three hidden layers with either eight or 16 hidden units per hidden layer.

For the results depicted in \cref{fig:results:conditioning:cartpole}~(C), we used FF agent architectures with one hidden layer with four hidden units and \textit{ReLU} activation function, $f_{\mathrm{ReLU}}(x)=\max(x, 0)$.

For the results depicted in \cref{fig:results:conditioning:cartpole}~(E), we used an RGRN agent architecture with one hidden layer with a single hidden neuron (see Appendix A in Ref.~\citenum{hartl2024evolutionary} for details).

\section{Cart-Pole Conditioning}
\label{app:agents:cartpole:conditioning}
In practice, we conditionally train the DM during \CHARLES optimization in \cref{sub:results:conditional:phenotype} jointly on the ANN parameters $\genome_i$ and the associated resting position ${x}^{(r)}_i$, the resting velocity ${\dot x}^{(r)}_i$, and the associated fitness score $f_i$ averaged over \params{$N_e=16$} episodes. 
Thus, the vector-valued conditions for given ANN parameters $\genome_i$ comprises $\condition^{(r)}_i=\{{x}^{(r)}_i, {\dot x}^{(r)}_i, f_i\}$.



\end{document}